\documentclass[a4paper,fleqn]{cas-dc}

\usepackage[authoryear]{natbib}

\usepackage{lineno}

\usepackage{todonotes}
\usepackage{algorithm}
\usepackage{algpseudocode}
\usepackage{tabularx}
\usepackage{multirow} 
\usepackage{booktabs} 

\usepackage[dvipsnames]{xcolor}
\usepackage{soul}

\definecolor{lightblue}{RGB}{173,216,230} 
\newcommand{\hlblue}[1]{{\sethlcolor{lightblue}\hl{#1}}}

\DeclareRobustCommand{\hlyellow}[1]{{\sethlcolor{yellow!75}\hl{#1}}}

\let\printorcid\relax 

\def\tsc#1{\csdef{#1}{\textsc{\lowercase{#1}}\xspace}}
\tsc{WGM}
\tsc{QE}
\tsc{EP}
\tsc{PMS}
\tsc{BEC}
\tsc{DE}

\begin{document}
\let\WriteBookmarks\relax
\def\floatpagepagefraction{1}
\def\textpagefraction{.001}
\shorttitle{}
\shortauthors{Freeman et~al.}

\title [mode = title]{Autonomous Apple Fruitlet Sizing and Growth Rate Tracking using Computer Vision}                      



\author[1]{Harry Freeman} 
\cormark[1]
\ead{hfreeman@cs.cmu.edu}
\author[1]{Mohamad Qadri} 
\ead{mqadri@cs.cmu.edu}
\author[1]{Abhisesh Silwal} 
\ead{asilwal@cs.cmu.edu}
\author[2]{Paul O'Connor} 
\ead{proconnor@umass.edu}
\author[1]{Zachary Rubinstein} 
\ead{zbr@cs.cmu.edu}
\author[2]{Daniel Cooley} 
\ead{dcooley@umass.edu}
\author[1]{George Kantor} 
\ead{gkantor@cs.cmu.edu}
\affiliation[1]{organization={Carnegie Mellon University, Robotics Institute},
            addressline={5000 Forbes Ave}, 
            city={Pittsburgh},
            postcode={15213}, 
            state={PA},
            country={USA}}
\affiliation[2]{organization={University of Massachusetts Amherst Stockbridge School of Agriculture},
            addressline={161 Holdsworth Way}, 
            city={Amherst},
            postcode={01003}, 
            state={MA},
            country={USA}}
\cortext[cor1]{Corresponding author}


\begin{abstract}
In this paper, we present a computer vision-based approach to measure the sizes and growth rates of apple fruitlets. Measuring the growth rates of apple fruitlets is important because it enables apple growers to determine when to apply chemical thinners to their crops in order to optimize yield. The current practice of obtaining growth rates involves using calipers to manually record the sizes of fruitlets over multiple days. Due to the number of fruitlets that need to be sized, this method is laborious, time-consuming, and prone to human error. With images collected by a handheld stereo camera, our system segments, clusters, and fits ellipses to fruitlets to measure their diameters. The growth rates are then calculated by temporally associating clustered fruitlets across days. We provide quantitative results on data collected in an apple orchard for Fuji, Gala, and Honeycrisp varietals across the 2021 and 2022 growing seasons. We demonstrate that our system is able to predict abscission rates within 4\% of the current manual method, 
while requiring significantly less manual effort. Moreover, we provide results on images captured by a robotic system in the field and discuss the next steps required to make the process fully autonomous.
\end{abstract}




\maketitle

\section{Introduction}\label{introduction}
Recent advancements in computer vision have allowed farmers to deploy autonomous plant monitoring solutions to more efficiently inspect vast quantities of crops. Computer vision-based systems provide fast and reliable information for downstream tasks such as harvesting (\cite{harvesting_new}), phenotyping (\cite{phenotyping_new}), and yield prediction (\cite{yield_new}), ultimately enabling farmers to make real-time crop management decisions. Common computer vision applications in agriculture include mapping (\cite{11077662}), counting (\cite{freman_sorghum}), modeling (\cite{modelling_new}), and disease classification (\cite{disease_class_new}).

We focus on the target application of measuring the growth rates of apple fruitlets%
{\renewcommand\thefootnote{\arabic{footnote}}\footnote{A fruitlet is a young apple formed on the tree shortly after pollination.}}%
, which is important because it enables farmers to better control their annual yield.  It is standard practice to thin apple trees to prevent them from developing a pattern of alternate year bearing (\cite{alternate_bearing}) in order to produce a more consistent yearly harvest.  To predict the effect of thinning application on trees, the Fruitlet Growth Model developed by \cite{greene2013development} is used to determine how often farmers need to spray their crops. The model takes into account the growth rates of a subset of fruitlets over multiple days.  The fruitlets with growth rates greater than 50\% of the fastest growing fruits are predicted to persist the thinning. Farmers calculate the percentage of fruitlets expected to abscise and use this information to determine if another round of chemical thinner needs to be applied. 

The sizing method most commonly used in practice involves identifying each individual fruitlet, using a digital caliper to hand-measure sizes%
{\renewcommand\thefootnote{\arabic{footnote}}\footnote{Throughout this work, size refers to the apparent diameter of a fruitlet measured along its minor axis in the 2D image plane.}}%
, and manually entering the data into a spreadsheet so that growth rates can be tracked. We will refer to this sizing process as the caliper method. The number of fruit typically sized using the caliper method is approximately 400-500 fruitlets per varietal. This means growers need to size hundreds to thousands of fruitlets depending on the number of varietals that they grow. Each fruitlet is sized twice per thinning application: once three to four days after application, and again seven to eight days after application. Taking this many hand-measurements is not only labor-intensive, but highly subject to human error. It is inefficient and time-consuming for a human to record caliper readings for hundreds to thousands of fruitlets across multiple days, making some farmers hesitant to adopt the approach. Manually associating fruitlets across different days is also very challenging; fruitlets are likely to have moved or fallen off, resulting in them being misidentified, which negatively affects growth estimates. Moreover, using calipers creates variability when measuring asymmetrically shaped fruit, leading to inaccurate measurements, which become more pronounced as different workers are employed to collect data. As a result, there is a need to automate this process to make sizing and temporal association faster, more repeatable, and more accurate.

There has been significant work dedicated towards both detecting and sizing fruit in agriculture. Studies have focused on pre-harvest detection and segmentation for yield and crop load estimation of apples, employing classical methods based on shape and color features (\cite{GONGAL201626}), deep-learning approaches (\cite{Bargoti, CHU2021206, TIAN2019417}), and Gaussian Mixture Models (\cite{ROY2019104897}). However, these works primarily focus on detecting and estimating the yield of fully developed apples, rather than on the sizing or growth tracking of smaller, developing fruit. For apple fruitlet detection, \cite{WANG2021271} developed a deep-learning-based approach utilizing YOLOv5 (\cite{YOLO_BASE}) and channel-pruning. Their work is dedicated towards only detection and not clustering or sizing. Similarly, \cite{KHANAL20238914} presented a YOLO-based method for fruitlet detection and cluster the fruitlets using a K-means approach. However, their algorithm requires both the maximum number of clusters and the maximum number of items per cluster to be specified in advance.

For fruit sizing, previous works have explored using reference objects for scale estimation. In the work of \cite{yield_1}, calibration spheres are placed on trees and used as reference scales to estimate the sizes of segmented apples. Similarly, \cite{smart_phone} sized fruits in the field with a smartphone by placing a reference circle of known size behind the fruit. While these methods only require simple segmentation and sizing algorithms, they do not extend well to fruitlets as it is impractical to place reference objects behind hundreds of fruit in occluded environments.

Approaches have also been developed to size fruit in 3D. Reconstruction-based methods are used by \cite{recon_0} and \cite{recon_1}, where 3D models are created from multiple sensor measurements. However, these methods are computationally expensive and do not perform well with occlusions where reconstructions are often incomplete. To address these issues, automated shape completion methods have been implemented which fit superellipsoids to accumulated (\cite{sweet_pepper, superellipsoid}) or partially accumulated (\cite{zaenker-new}) point clouds. 
These methods either rely on successive frame alignment algorithms which fail in agricultural environments due to the dynamic structure of plants, or use expensive ray casting operations to maintain 3D maps which are slow when performed at finer resolutions. Learning-based 3D shape completion methods have also been recently presented (\cite{shape-0, shape-1, modelling_new}), but the fruits are much larger, making it easier to capture enough of the fruit surface for shape completion. \cite{gongal} performed 3D sizing, where the major-axis of an apple is fit to 3D points collected from a single camera image and time-of-flight sensor. However, the performance is poor, achieving an accuracy of 69.1$\%$. The work of \cite{Photogrammetric-apple} also sizes apples from single images by fitting spheres in 3D. This would not adapt well to apple fruitlets due to their small size making it challenging to fit 3D geometric shapes from single images. More similar to our work, ~\cite{Sapkota} detected apple fruitlets utilizing a Yolov8 network and evaluated the performance of different 2D and 3D shape fitting techniques to size them.

As an alternative to 3D sizing, 2D photogrammetric methods have been adopted in agriculture. These methods either directly estimate the widths of fruits (\cite{gongal}) or fit ellipses to measure the heights and widths (\cite{on_tree, olive}). This is advantageous as sizes are able to be extracted quickly from 2D images without the need to aggregate information from multiple views. However, the presented approaches use simple detection and segmentation modules, relying on color information and non-deep features.  While these methods work well in their respective domains, they would fail when trying to detect and segment fruitlets. This is because fruitlets are much closer together, and their colors blend in with the surrounding leaves. Most similar to our work, \cite{Qadri} fitted ellipses to apple fruitlets to estimate their diameters, while \cite{Checola}, in concurrent work, lifted detected 2D bounding boxes to 3D in order to size fruitlets in extracted high-quality images from video sequences. Unlike these studies, our method associates fruitlets across days to predict abscission rates.

There have been works dedicated towards  spatio-temporal registration in agriculture. 4D association methods have been proposed at whole plot scales for large rows of crops ({\cite{field_0, field_1, field_2, field_3, field_4}). These methods assume static scenes during data collection and take advantage of the structural similarity of rows throughout the growing season. This does not apply when imaging apple fruitlets as the local structure of the scene is more likely to change when images are taken up close. Approaches have also been proposed for the temporal association of individual plant components on a single stem, utilizing different methods such as (skeletonization \cite{skel_0, skel_1, skel_2}), self-organizing maps (\cite{som}), and fitting statistical models (\cite{ssm}). However, the datasets used  were collected in controlled lab environments using a high-precision laser scanner on a single isolated plant. It would be challenging to extend these methods towards temporally associating apple fruitlets in the field due to occlusions, incorrect detections, and incomplete point clouds. \cite{fixed_apple} developed a deep-learning approach to size apples and track growth rates over time. However, images are captured from a camera fixed in place, and fruit detection is performed by comparing center coordinates of segmented apples. This does not adapt well to fruitlets as a result of their small size, close proximity, and number of fruit needed to be sized.
\cite{rel_row} proposed a method for spatio-temporal matching of multiple plant point clouds using RGB-D images. They later extended their work to deformably register a map to new images (\cite{10802321}) and map fruit growth (\cite{11112569}). Their approach does not fit our use case, as they utilized the assumption that structurally similar images were captured by a robot driving down a row of crops. This is demonstrated through the use of visual place recognition descriptors and the use of plant stems as temporally stable features. \cite{11077662} utilized LiDAR, RGB, and IMU measurements from a robot driving down orchard rows to build semantic point clouds and spatio-temporally track fruit through frame alignment and matching.
Transformer-based methods have also been proposed for spatio-temporal matching (\cite{reidentify, transformer_fruitlet}), but they require large-scale annotated datasets for training, making them difficult to apply in agricultural settings where labeled data is scarce and expensive to collect.

\begin{figure*}
	\centering
	\includegraphics[width=.9\textwidth]{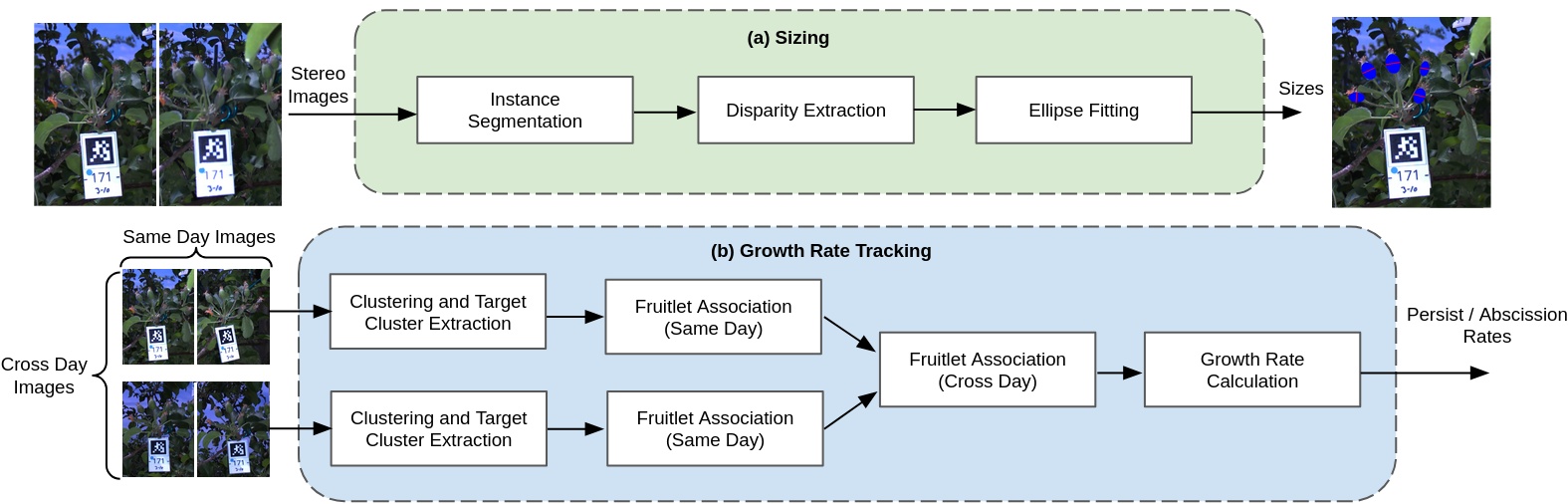}
	\caption{Overview of our (a) fruitlet sizing and (b) growth rate tracking pipelines used to predict persist and abscission rates.}
    \label{fig:full_pipeline}
    \vspace{-15pt}
\end{figure*}

\cite{riccardi} presented a fruit tracking system for the temporal registration of strawberries. They introduced a histogram-based descriptor that captures the relationship of a fruit's position to that of its neighbors, which could be applied to associating apple fruitlets. However, the descriptor is not invariant to \(\text{SE}(3)\) transformations, and as a result its performance is dependent on robust initial scene alignment. 
In their work, this is easier to achieve as data is captured by a robot driving down a row of crops with a high-precision laser scanner. In our system, initial scene alignment is more challenging, as close-up stereo images produce less similarly structured scenes and are more prone to occlusions and missed detections.

Despite substantial progress in fruit detection, sizing, spatio-temporal association, and yield estimation, current methods do not provide a complete, end-to-end solution for accurately measuring and tracking the growth rates of apple fruitlets over time to be used for reliable thinner response prediction. 
To address these gaps, this study aims to develop and evaluate a computer vision–based system capable of autonomously measuring apple fruitlet sizes and tracking growth rates across days using stereo image pairs collected in a commercial orchard. The main contributions and research objectives are fourfold:
\begin{enumerate}
\item[i] To design a vision pipeline that can robustly size, cluster, and temporally associate apple fruitlets in orchard environments for automated thinner response prediction.
\item[ii] To quantitatively evaluate the accuracy of computer vision–based thinner response prediction compared to those obtained using traditional handheld caliper measurements.
\item[iii] To investigate the effectiveness of individual pipeline components through ablation studies and method comparisons.
\item[iv] To demonstrate the adaptability of the proposed approach by conducting experiments with both a handheld stereo camera and a robotic system in a commercial apple orchard.
\end{enumerate}

We demonstrate that our system produces results comparable to those obtained using the caliper method, with the ability to reduce the labor-intensive effort and significantly improve speed.

\section{Materials and Methods}
~\label{sec:methodology}
\vspace{-20pt}
\subsection{System Overview}~\label{sec:system_overview}
The Fruitlet Growth Model presented by \cite{greene2013development} specifies that growers select the fruitlet clusters\footnote[3]{A cluster is a group of fruitlets that grow out of the same bud.} they want to size and measure the growth rates of all fruitlets in each cluster. Our system is designed to replicate this process by both sizing the fruitlets and autonomously identifying which fruitlets in the image belong to the cluster intended to be sized. We refer to this cluster as the target cluster.

Images are captured in the field using an illumination-invariant flash stereo camera. Once images of all clusters are acquired, the data is passed through our sizing and growth rate tracking pipelines (Fig.~\ref{fig:full_pipeline}). In the first stage, 
the fruitlets are detected and segmented
using a Mask-RCNN (\cite{mask-rcnn}) instance segmentation network, and disparities are extracted using RAFT-Stereo (\cite{lipson2021raft}) to build point clouds of each fruitlet. Ellipses are fit to the fruitlets and the diameter of each fruitlet is calculated using the minor axis of the fit ellipse and the disparity values. 

The second stage consists of clustering, temporal fruit association, and growth rate tracking. The extracted point clouds are used to cluster the fruitlets using a multi-stage Louvain Community Detection (\cite{community_cluster}) approach, and the fruitlets corresponding to the target cluster are then identified. Fruitlet clusters in images taken on the same and on different days are associated using an approach that makes use of Iterative Closest Point (ICP) and the Hungarian Algorithm (\cite{hungarian}). The sizes of the resulting temporally associated fruitlets are used to calculate growth rates, which in turn are used to predict the percentage of fruitlets that will persist and abscise.
\subsection{Fruitlet Cluster Sampling}~\label{sec:cluster_sample}
The Fruitlet Growth Model specifies that growers choose the number of trees and the number of clusters per tree they want to size. While the original model suggests using 15 clusters per tree across 7 trees for a total of 105 clusters, we instead use 7 clusters per tree across 10 trees for a total of 70 clusters. We use fewer clusters because this practice has been previously adopted by apple growers to save time acquiring data. For example, Michigan State University\footnote[4]{\url{https://mediaspace.msu.edu/media/How+to+Use+the+Fruit+Growth+Model+to+Guide+Apple+Thinning+Decisions/1_o359pflp}} and the University of Massachusetts\footnote[5]{\url{https://ag.umass.edu/fruit/fact-sheets/hrt-recipe-predicting-fruit-set-using-fruitlet-growth-rate-model}} suggest using 5 trees with either 15 or 14 clusters per tree respectively. Additionally, spreading clusters over more trees allows us to collect data on a wider range of fruitlet sizes and appearances. This is advantageous as it allows us to evaluate our system's performance across a more diverse and representative set of fruitlets. Because our objective is to compare predicted thinner response between computer vision and hand-caliper growth rates, and not compare against actual abscission rates, this is a fair sampling strategy for the purpose of this work.

We hang AprilTags (\cite{april_tag}) to identify the selected clusters and fruitlets. AprilTags were selected because they allow for fast identification in computer vision systems. The AprilTag is only used for identifying the cluster id in addition to determining which fruitlet cluster is the target cluster, as described in Sec.~\ref{sec:fruitlet_cluster}. Each fruitlet in each cluster is assigned a unique id written on the back which is used for evaluating against ground-truth. An example of tagged fruitlet trees and clusters can be seen in Fig. \ref{fig:cluster_example}.

\begin{figure}
	\centering
	\includegraphics[width=0.99\columnwidth]{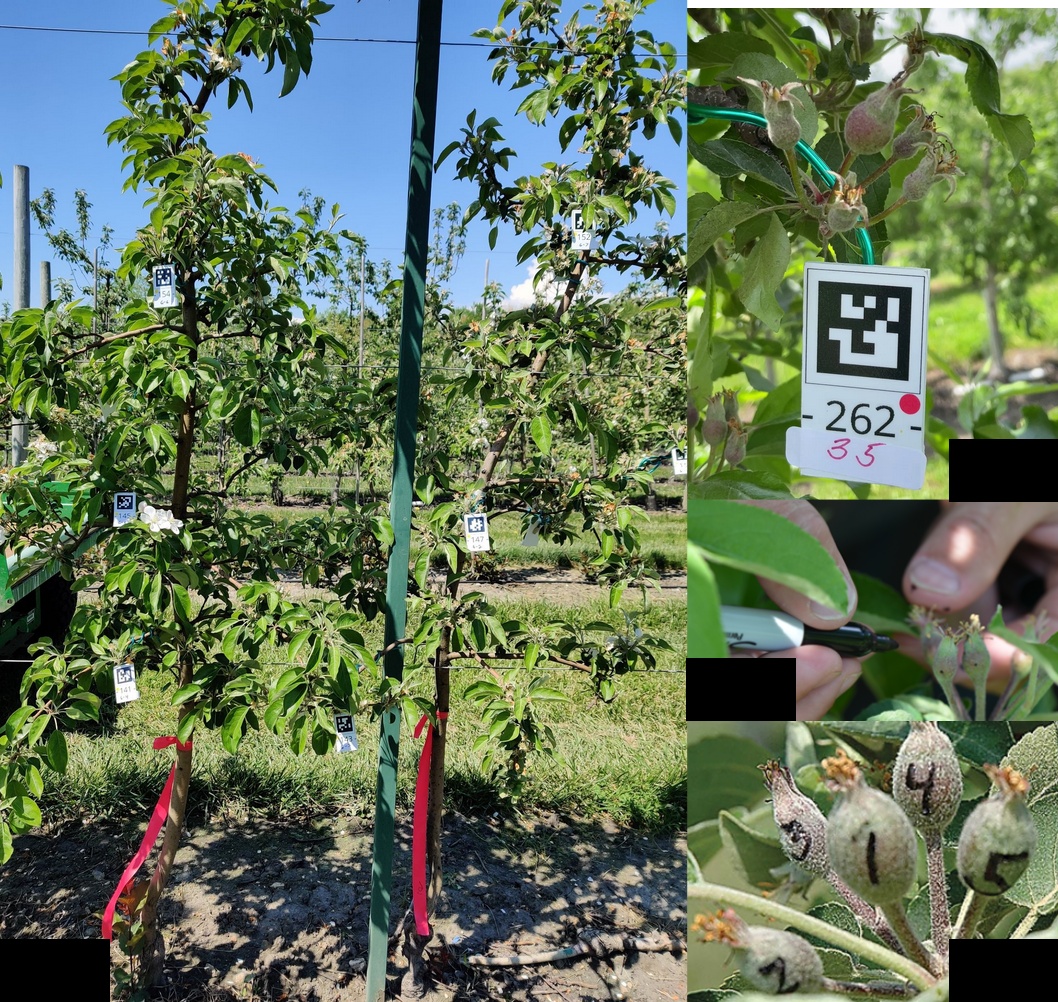}
    \put(-17, 4){\textcolor{white}{\small{(d)}}}
\put(-17, 115){\textcolor{white}{\small{(b)}}}
\put(-229, 5){\textcolor{white}{\small{(a)}}}
\put(-76, 67){\textcolor{white}{\small{(c)}}}
	\caption{(a) Example of tagged fruitlet trees and clusters. (b) An AprilTag hung next to the cluster. (c, d) Individual fruitlets, each assigned a unique ID written on the back for identification.}
    \label{fig:cluster_example}
    \vspace{-15pt}
\end{figure}

\subsection{Image Capture}~\label{sec:image_capture}
Images are captured using an in-hand version of the illumination-invariant flash stereo camera presented by \cite{9636542}. Stereo camera systems are commonly used in agriculture to extract 3D information as a result of their reliability and low cost. They resolve finer details compared to systems that use LiDAR and flash-based systems are more resilient to varying illumination conditions (Fig.~\ref{fig:inhand_camera} c-d) where RGB-D sensors inconsistently perform.

The imaging system consists of a compact active flash stereo camera. The design follows the flash stereo principle described by \cite{9636542}, in which high-power pulsed illumination reduces the influence of ambient outdoor lighting and enables consistent image formation under varying illumination conditions. The system integrates two synchronized color cameras arranged in a fixed stereo configuration with a 0.06m baseline and an integrated high-power LED flash. It is implemented in a smaller and lighter form factor than the one presented by ~\cite{9636542} to enable handheld operation and robotic mounting, using smaller stereo modules and modified optics while preserving the same active stereo imaging. The system operates over a range of approximately 0.2–1.5m. The hardware specifications are summarized in Table \ref{table:camera_specs}.

\begin{table}[h]
\centering
\caption{Hardware specifications of the active flash stereo camera.}
\begin{tabular}{ll}
\hline
Sensor & Sony IMX296 (color) \\
Resolution & 1.6 megapixels \\
Lens focal length & 6 mm \\
Field of View (FoV) & 60$^\circ$ $\times$ 45.4$^\circ$ \\
Stereo baseline & 0.06m \\
Operational range & 0.2--1.5m \\
Illumination & 500 W CREE LED \\
Flash color temperature & 5600 K \\
\hline
\end{tabular}
\label{table:camera_specs}
\vspace{-10pt}
\end{table}

To facilitate data collection, we designed a custom setup that consists of the stereo camera connected via USB to a laptop to save the captured images. A phone is mounted on the back and connected via USB-C to the same laptop to allow the user to visualize and assess the quality of the images in real-time. When imaging, for each cluster on each day of data collection, an unoccluded image of the corresponding AprilTag was first captured, followed by one or two images of the fruitlet cluster.

An unoccluded image of the AprilTag was captured first to ensure reliable identification of the target cluster. This is for cases where the tag may be partially occluded or not fully visible when imaging the full cluster, which could prevent successful detection by the AprilTag software, as further discussed in Sec.~\ref{subsec:april_tag}. The reason for capturing only one to two images per cluster, as opposed to longer video sequences (\cite{Checola}), is that we want to limit both the amount of time apple growers need to spend in the field acquiring images and the computational time required to post-process the data. Taking many images or long video sequences significantly adds time to both processes, which results in reduced benefit. Ideally, one image per cluster is taken, but two images are used when not every fruitlet is visible in a single image as a result of occlusions. As we demonstrate in Sec.~\ref{subsec:sizing_results}, capturing a second image improves sizing and growth rate tracking by helping mitigate the effect of occlusions. However, it does not fully resolve all cases, particularly in clusters with severe overlap or dense surrounding leaves. During image acquisition, the operator aimed to maintain a distance of approximately 40cm between the stereo camera and the target cluster. Because imaging was performed manually under varying field conditions, the distances ranged from roughly 20cm to 75cm.

\begin{figure}
	\centering
	\includegraphics[width=0.99\columnwidth]{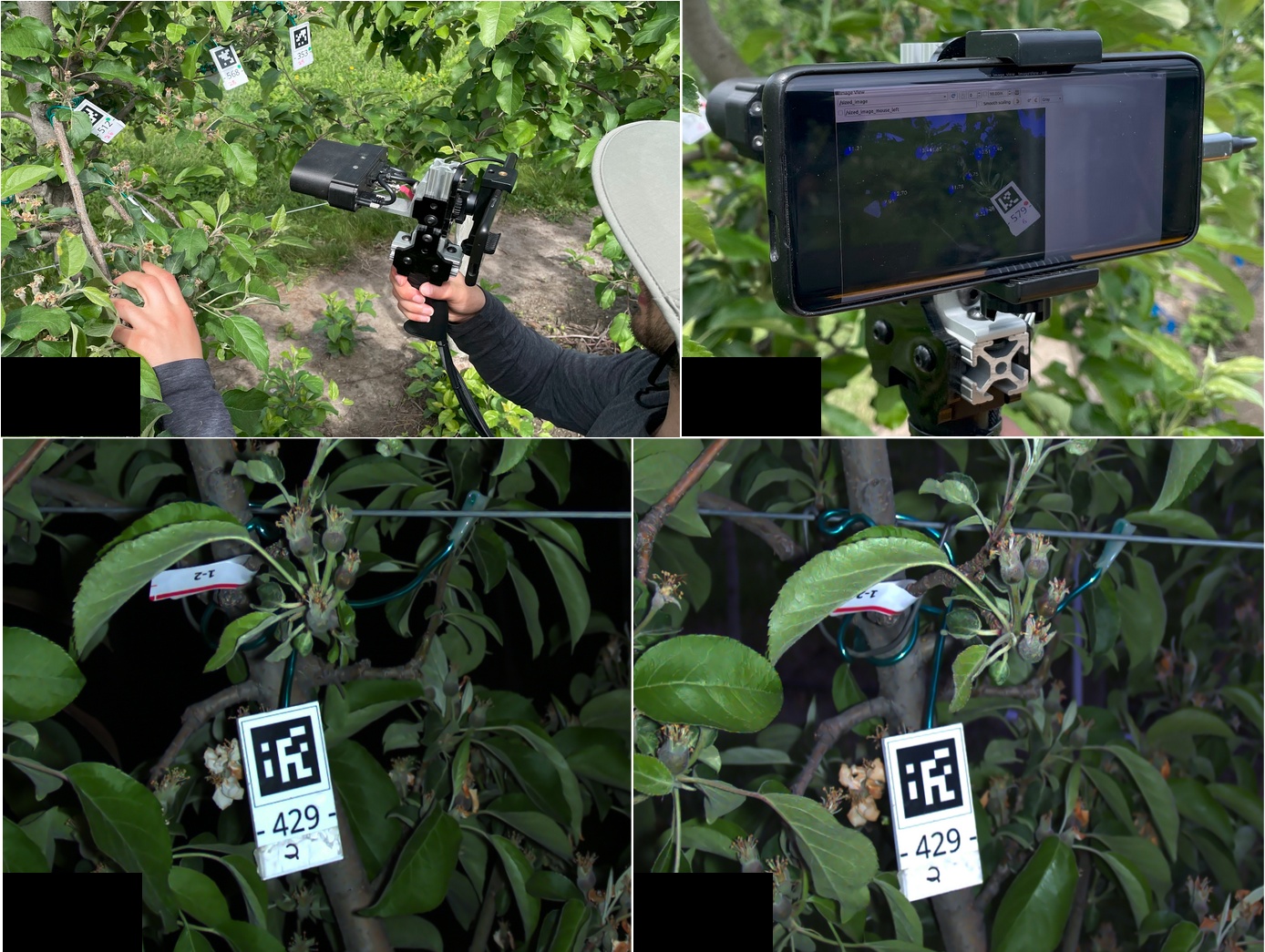}
    \put(-228, 101){\textcolor{white}{\small{(a)}}}
    \put(-102, 101){\textcolor{white}{\small{(b)}}}
    \put(-228, 5){\textcolor{white}{\small{(c)}}}
    \put(-110, 5){\textcolor{white}{\small{(d)}}}
	\caption{(a) Handheld illumination-invariant flash stereo camera. (b) Phone mounted and connected via USB-C to display images to the user in real time. (c) Image captured at midnight in the dark. (d) Image captured at noon in bright daylight, appearing similar due to the flash stereo camera.}
	\label{fig:inhand_camera}
    \vspace{-18pt}
\end{figure}

\subsection{Segmentation and Disparity Estimation}~\label{sec:seg_disp}
\vspace{-10pt}

Two differences between our approach and similar works presented in Sec.~\ref{introduction} are that we use RAFT-Stereo and Mask-RCNN for disparity estimation and semantic segmentation respectively. RAFT-Stereo is a deep-learning-based stereo matching network that outperforms traditional disparity generation methods, such as SGBM \cite{SGBM_2008}, on benchmarks such as the Middlebury Stereo Evaluation\footnote[6]{\url{https://vision.middlebury.edu/stereo/eval3/}}. RAFT-Stereo does not require fine-tuning, which is advantageous as the network does not need to be retrained between datasets or when switching out the stereo camera being used. As well, the method produces results comparable to state-of-the-art disparity estimation networks (\cite{wen2025foundationstereozeroshotstereomatching}) when run in the field (Sec.~\ref{subsec:sizing_results}). 

To account for the difficulties in obtaining ground-truth segmented data, we use a Mask-RCNN-based framework instead of YOLO and pix2pix (\cite{pix2pix})-based methods used in related works (Sec.~\ref{introduction}), as it allows us to utilize a two-stage training process that reduces labeling effort. In the first stage, bounding box annotations are used to train the bounding box regression and classification heads. In the second stage, the weights of these heads are frozen, and a subset of the fruitlets in each image are segmented with Segment Anything (\cite{segmentanything}) and used to train the mask head. An overview of the process can be seen in Fig.~\ref{fig:example_det_and_seg}. We use a customized detectron2 (\cite{wu2019detectron2} ) Mask-RCNN implementation.

\begin{figure}
	\centering
	\includegraphics[width=0.99\columnwidth]{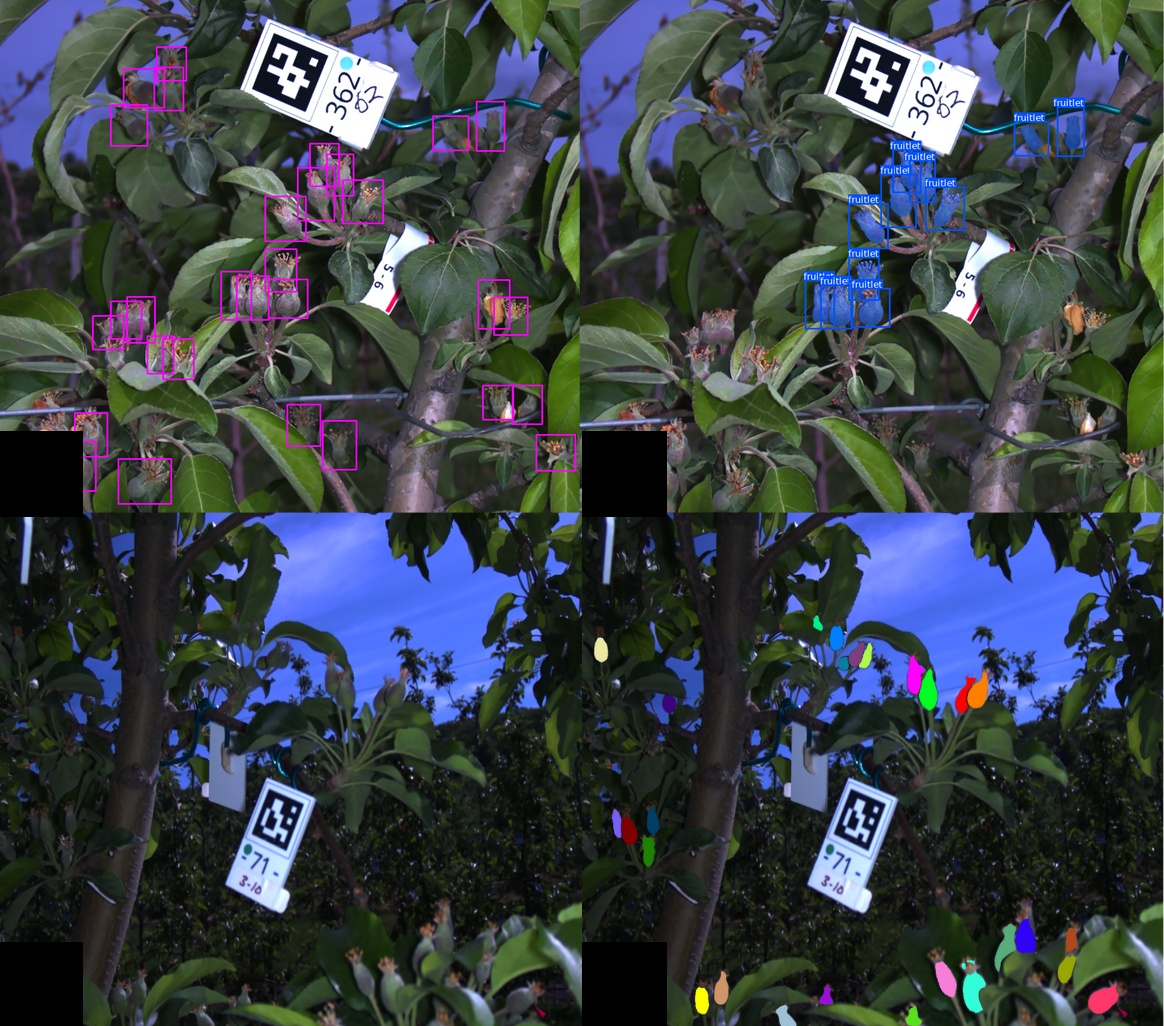}
    \put(-233, 109){\textcolor{white}{(a)}}
    \put(-115, 109){\textcolor{white}{(b)}}
    \put(-233, 5){\textcolor{white}{(c)}}
    \put(-115, 5){\textcolor{white}{(d)}}
    \caption{(a) Bounding box annotations around each fruitlet used to train the Mask-RCNN bounding box regression and classification heads. (b) A subset of fruitlets segmented and used to train the mask head. (c–d) Example input image and corresponding output of the final trained Mask-RCNN network.}
    \label{fig:example_det_and_seg}
    \vspace{-20pt}
    \end{figure}

\subsection{Ellipse Fitting and Sizing}
An ellipse is fit to the segmented fruitlets following the process demonstrated in Fig. \ref{fig:example_ellipse_fit}.  First, for each fruitlet, the contour surrounding the segmented points is extracted. Then, an ellipse is fit using the OpenCV (\cite{opencv_library}) fitEllipse function. Finally, the size is calculated following Equation~\ref{eq:size}, where $b$ is the baseline of the stereo camera, $d_{\text{med}}$ is the median disparity value of all segmented fruitlet pixels (Fig. \ref{fig:example_ellipse_fit}(b)) with valid disparity measurements, and $a_{\text{minor}}$ is the length of the minor axis of the fitted ellipse. 
\begin{equation}
    \text{\quad \quad \quad \quad size} = \frac{b \times a_{\text{minor}}}{d_{\text{med}}}
    \label{eq:size}
\end{equation}

\begin{figure}
	\centering
	\includegraphics[width=0.95\columnwidth]{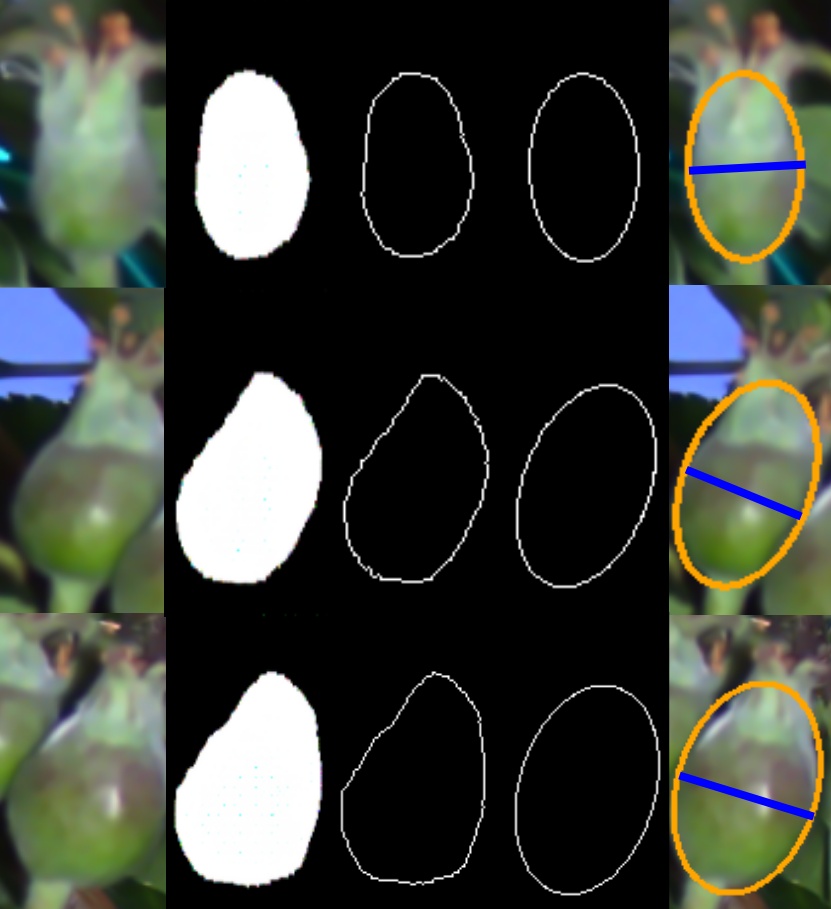}
    \put(-207, 252){{(a)}}
     \put(-165, 252){{(b)}}
     \put(-120, 252){{(c)}}
     \put(-74, 252){{(d)}}
     \put(-28, 252){{(e)}}
     \caption{Fruitlet ellipse fitting example. (a) Input image of a fruitlet. (b) Mask-RCNN segmentation output. (c) Contour fitted to the segmentation mask. (d) Ellipse fitted around the contour. (e) Fruitlet diameter computed from the fitted ellipse, indicated by the blue line.}
    \label{fig:example_ellipse_fit}
    \vspace{-20pt}
    \end{figure}

\subsection{Fruitlet Clustering}~\label{sec:fruitlet_cluster}
In order to measure and compare the growth rates of the fruitlets, our system needs to be able to identify which fruitlets belong to the target cluster. We achieve this using a multi-stage Louvain Community Detection approach. Louvain Community Detection is an algorithm that extracts communities within a graph using a heuristic method based on modularity optimization. A community is a subset of nodes whose connections among themselves are denser than connections with the rest of the graph. The algorithm iterates through nodes and moves each node into the community that maximizes modularity gain. The process repeats for each node until no modularity gain can be found. We use the NetworkX (\cite{networkx}) Louvain Community Detection implementation.

Point clouds of each fruitlet are first extracted using the segmentations and disparities from Sec.~\ref{sec:seg_disp}. The point clouds are filtered using a bilateral filter and a depth discontinuity filter (\cite{6906903}). The resulting number of valid depth values varies per fruitlet between approximately 50 and 2500, with smaller and more occluded fruitlets generally yielding fewer points due to limited visible surface area. Point clouds with fewer than 100 valid points are discarded.

Our clustering approach is then performed as described in Algorithm~\ref{alg:cluster_detect}. A graph $G_1$ is first constructed and consists of the centroids $c_{i} \in \mathbb{R}^3$ of each fruitlet $i \in \{1, ..., N\}$ as a node~(Line 1, Algorithm~\ref{alg:cluster_detect}). An edge $e_{ij}$ is created between nodes if their Euclidean distance is less than a pre-determined threshold $\tau_1$~(Line 2, Algorithm~\ref{alg:cluster_detect}). The weight $w_{ij}$ of edge $e_{ij}$ is determined based on the Euclidean distance, with closer nodes receiving a greater edge weight~(Line 3, Algorithm~\ref{alg:cluster_detect}). Louvain Community Detection is then run on $G_1$ to produce a set of communities $H_1$~(Lines 4-5, Algorithm~\ref{alg:cluster_detect}). 

As a result of the inconsistent spacings between adjacent fruitlets and fruitlets falling off during the thinning period, naively using only Louvain Community Detection often results in fruitlet clusters being incorrectly divided into multiple communities (Fig.~\ref{fig:example_cluster} (b)). We observed that when this occurs one of the communities almost always contains two fruitlets. To account for this, we iterate through all communities of size two and combine them with their nearest communities if the distance between the centroids of all the fruitlets in the combined community and the centroid of the combined community is less than a threshold $d_c$~(Line 6, Algorithm~\ref{alg:cluster_detect}).

\begin{figure}
\begin{algorithm}[H]
    \textbf{Input}: $C = \{c_{i} \forall \ i \in \{1, ..., N\}\}$\\
    \textbf{Parameters}: $\tau_1$, $\tau_2$, $d_c$, $d_u$\\   
    \textbf{Output}: $H$
    \begin{algorithmic}[1]
        \State $V_1 \gets C$
        \State $E_1 \gets \{e_{ij} \forall \ i, j \  \text{s.t.} \ \lVert v_i - v_j \rVert < \tau_1: v_i \in V_1, v_j \in V_1\}$
        \State $W_1 \gets \{w_{ij} = 1-\frac{\lVert v_i - v_j \rVert}{\tau_1} \forall \ i, j \in E_1\}$
        \State $G_1 \gets (V_1, E_1, W_1)$
        \State $H_1 \gets $ LCD($G_1$)
        \State $H_1 \gets $ MergeCommunities($H_1$, $d_c$)
        \State $V_2 \gets C \backslash \cup_{h \in H_1} $
        \State $E_2 \gets \{e_{ij} \forall \ i, j \  \text{s.t.} \ \lVert v_i - v_j \rVert < \tau_2, v_i \in V_2, v_j \in V_2\}$
        \State $W_2 \gets \{w_{ij} = 1-\frac{\lVert v_i - v_j \rVert}{\tau_2} \forall \ i, j \in E_2\}$
        \State $G_2 \gets (V_2, E_2, W_2)$
        \State $H_2 \gets $ LCD($G_2$)
        \State $H \gets H_1 \cup H_2$
        \State $H \gets $ MergeCommunities($H$, $d_c$)
        \State $H \gets $ MergeUnassigned($H$, $C$, $d_u$)
    \end{algorithmic}
    \caption{Fruitlet Clustering}
    \label{alg:cluster_detect}
\end{algorithm}
\vspace{-24pt}
\end{figure}

Occasionally, some clusters have all fruitlets unassigned to a community. To overcome this, the above process is repeated for all unassigned fruitlets using a relaxed threshold. A graph $G_2$ is built consisting of the unassigned fruitlet centroids, and edges and weights are created as before with a new threshold $\tau_2 > \tau_1$~(Lines 7-10, Algorithm~\ref{alg:cluster_detect}). Louvain Community Detection is run on this graph to produce a new set of communities $H_2$, which is combined with $H_1$ to form $H$~(Lines 11-12, Algorithm~\ref{alg:cluster_detect}). Once again, communities of size two are merged~(Line 13, Algorithm~\ref{alg:cluster_detect}). As a final step, all remaining unassigned fruitlets are merged with their nearest communities~(Line 14, Algorithm~\ref{alg:cluster_detect}) if the distance between the centroids of all the fruitlets in the combined community and the centroid of the combined community is less than $d_u$. A visualization of the final result can be seen in Fig.~\ref{fig:example_cluster} (c). 
Fruitlets that belong to the community closest to the AprilTag are considered part of the target cluster and are used for growth rate tracking (Fig.~\ref{fig:example_cluster} (d)).

\begin{figure}
	\centering
	\includegraphics[width=0.99\columnwidth]{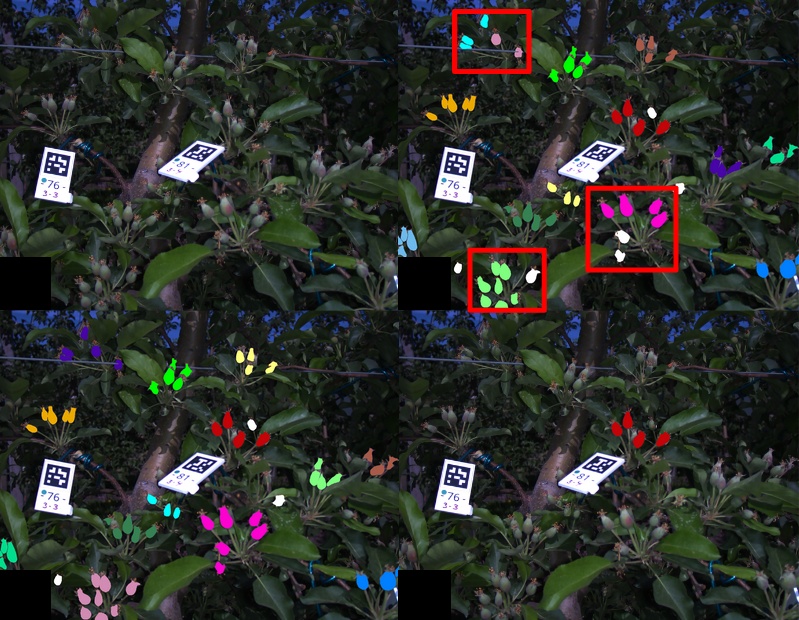}
    \put(-234, 97){\textcolor{white}{(a)}}
    \put(-116, 97){\textcolor{white}{(b)}}
    \put(-234, 5){\textcolor{white}{(c)}}
    \put(-116, 5){\textcolor{white}{(d)}}
    \caption{(a) Image with many fruitlet clusters. (b) Output of Louvain Community Detection. Red boxes highlight the same cluster being represented by multiple communities or unassigned fruitlets (white). (c) Fruitlet clustering output. (d) Fruitlets belonging to the community closest to the AprilTag are determined to belong to the target cluster.}
    \label{fig:example_cluster}
    \vspace{-20pt}
    \end{figure}

As discussed in Sec.~\ref{sec:clustering_performance}, multi-stage Louvain Community Detection outperforms other off-the-shelf clustering methods. We note, however, that DBSCAN (\cite{dbscan}) still generates competitive results despite having a worse clustering performance~(Table~\ref{table:cluster_pred}).

\subsection{{AprilTag Detection}}\label{subsec:april_tag}
One limitation of using AprilTags is that they can become partially occluded if not optimally placed. To mitigate this, the first image captured for each cluster is of the AprilTag itself, ensuring that the cluster can still be identified even if the tag is not fully visible in subsequent fruitlet images. Additionally, we augment the Mask-RCNN network to detect and segment AprilTags alongside fruitlets. In cases where the tag cannot be detected in the cluster images, its 3D position is estimated using the tag segmentations and disparity information. For images with more than one tag detection, the correct tag is identified by using a heuristic function that considers tag's depth relative to the camera as well as its 2D center position in the image. An overview of this process is illustrated in Figure~\ref{fig:april_tag_det}.

In the future, the tag instance segmentation step could be eliminated by placing multiple tags around each cluster to ensure that at least one tag remains visible under varying viewpoints and occlusion conditions.

\subsection{Temporal Association and Growth Rate Tracking}\label{sec:cross_day_association}

Association is performed using a coarse cluster alignment to fine fruitlet matching approach. First, for each image, we create a point cloud of the target cluster by combining the segmented point clouds of individual fruitlets. The cluster point clouds are aligned by finding the optimal transformation using ICP. The fruitlets are then associated using the Hungarian algorithm on the pairwise Euclidean distances of their transformed centroids. Matches whose Euclidean distance exceeds a specified threshold are discarded. This simple matching method provides reasonable performance to predict the persist and abscission percentages.

As demonstrated in Fig.~\ref{fig:full_pipeline} (b), the fruitlets are first clustered and associated in images taken on the same day, and the larger size of each fruitlet is used for calculating growth rates. If only one image is taken of a cluster on a particular day, the association step is skipped. Next, the fruitlets are temporally associated across days and growth rates are calculated by finding the difference between the end and start sizes.

Lastly, the fruitlets' response to the application of chemical thinner is predicted by calculating the persist and abscission rates following the process described in the Fruitlet Growth Model. This is achieved by determining the average growth rate of the top 20 fastest growing fruitlets. All fruitlets with growth rates greater than 50\% of this value are predicted to persist, and the rest abscise. We validate our calculations by comparing our results to those produced by the Predicting Fruitset spreadsheet available at the Michigan State University Extension Horticulture website\footnote[7]{\url{https://www.canr.msu.edu/apples/horticulture/}}.

\begin{figure}
\centering
\includegraphics[width=0.99\linewidth]{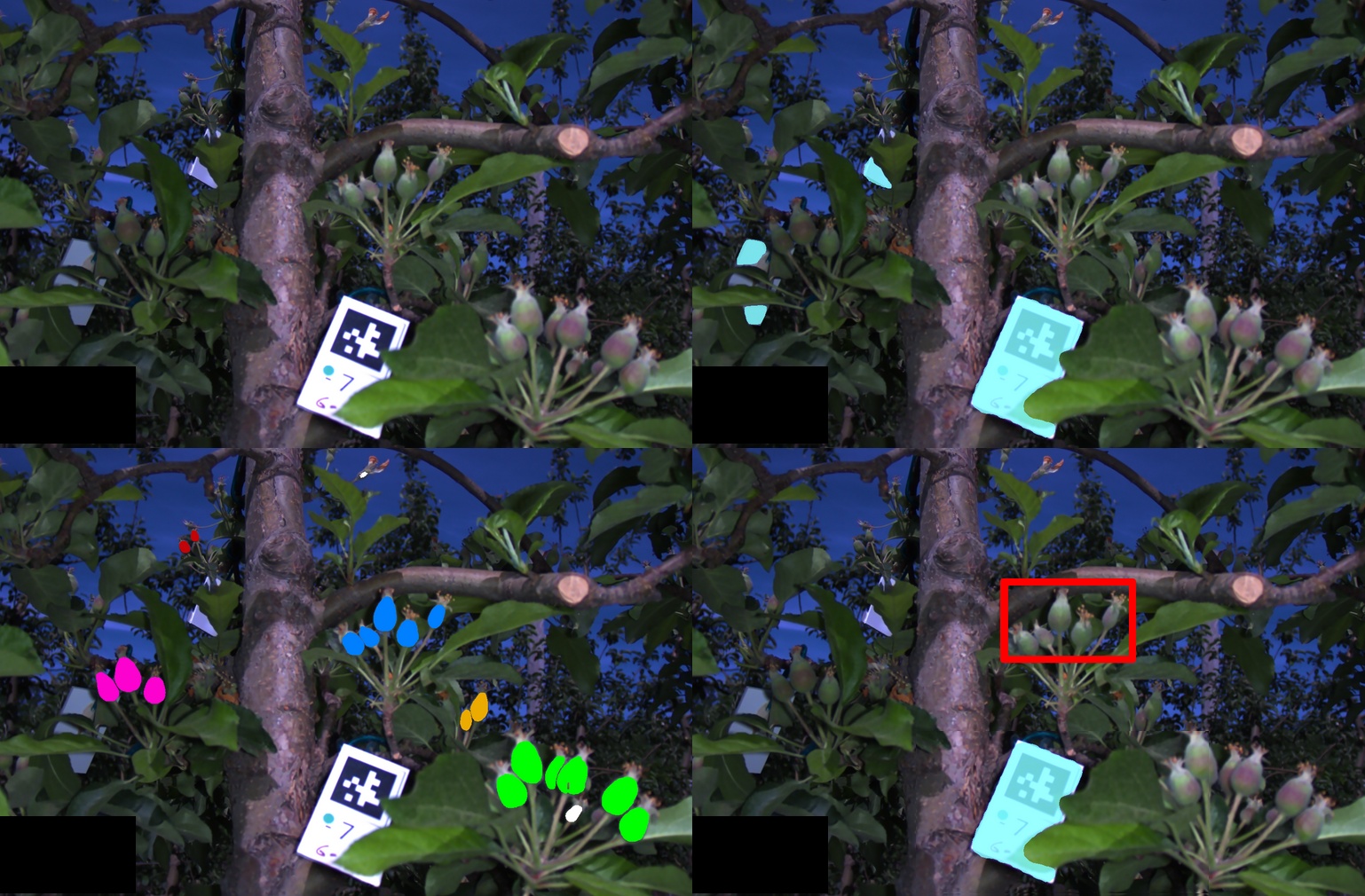}\put(-230, 82){\textcolor{white}{(a)}}
    \put(-110, 82){\textcolor{white}{(b)}}
    \put(-230, 5){\textcolor{white}{(c)}}
    \put(-110, 5){\textcolor{white}{(d)}}
\vspace{-10pt}
\caption{(a) Image with AprilTag partially occluded by a leaf which prevents detection. (b) Output of tag segmentation head from Mask-RCNN network with multiple tags detected. (c) Extracted fruitlet clusters. (d) The correct AprilTag and cluster is identified using a heuristic-based matching process.}
\label{fig:april_tag_det}
\vspace{-8pt}
\end{figure}

\begin{figure}
\centering
\includegraphics[width=0.99\linewidth]{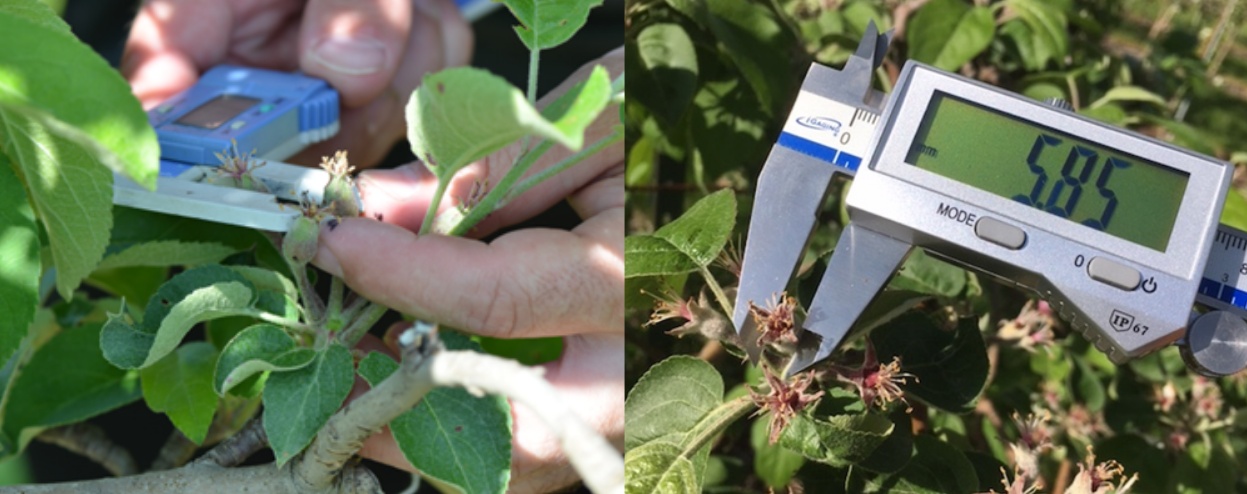}
\vspace{-15pt}
\caption{Example hand-caliper measurement of a fruitlet.}
\label{fig:caliper_example}
\vspace{-18pt}
\end{figure}

\subsection{Dataset}
\subsubsection{Data Collection}
\label{subsec:data_collection}
Our dataset~(\cite{freeman2026fruitlet}) consists of stereo images of 532 clusters along with their respective caliper measurements collected over three years. The data was collected at the University of Massachusetts Amherst Cold Spring Orchard. Each cluster was imaged three times over a period of five days, referred to as Day 1, Day 3, and Day 5. The breakdown of dates and varietals can be seen in Table~\ref{table:dataset_div}. Prior to data collection, the clusters were selected, tagged with an AprilTag, and the fruitlets were individually labeled following the process described in Sec.~\ref{sec:cluster_sample}. A human manually operated the handheld stereo camera (Fig. \ref{fig:inhand_camera}), collecting a sequence of images of each cluster. 
For evaluation, an image of the AprilTag followed by one to two images of each fruitlet cluster were manually selected per day, representing an apple grower capturing one to two photos of the same fruitlet cluster as described in Sec.~\ref{sec:image_capture}. Hand measurements were collected for each fruitlet using a digital caliper (Fig. \ref{fig:caliper_example}). Because measurements were taken for multiple varietals across different growing seasons, they were collected by multiple people.

\newcolumntype{A}{>{\centering\arraybackslash\hsize=.3\hsize}X}
\newcolumntype{B}{>{\centering\arraybackslash\hsize=.1\hsize}X}
\newcolumntype{D}{>{\centering\arraybackslash\hsize=.1\hsize}X}
\newcolumntype{E}{>{\centering\arraybackslash\hsize=.1\hsize}X}
\newcolumntype{F}{>{\centering\arraybackslash\hsize=.1\hsize}X}
\newcolumntype{G}{>{\centering\arraybackslash\hsize=.1\hsize}X}
\newcolumntype{H}{>{\centering\arraybackslash\hsize=.1\hsize}X}
\newcolumntype{I}{>{\centering\arraybackslash\hsize=.1\hsize}X}
\newcolumntype{J}{>{\centering\arraybackslash\hsize=.2\hsize}X}
\newcolumntype{K}{>{\centering\arraybackslash\hsize=.1\hsize}X}
\newcolumntype{Q}{>{\centering\arraybackslash\hsize=.05\hsize}X}

\begin{table*}[!t]
\scriptsize
\centering
\caption{Summary of fruitlet datasets collected over 3 years.}
\newcolumntype{Y}{>{\centering\arraybackslash}X}
\newcolumntype{L}{>{\raggedright\arraybackslash}X}
\begin{tabularx}{\linewidth}{@{}YYYYLYY@{}}
\hline \hline
\addlinespace[0.5ex]
Year & Day 1 & Day 3 & Day 5 & Varietal & Train Clusters & Eval Clusters \\
\hline
\multirow{3}{*}{2021} & \multirow{3}{*}{05/21} & \multirow{3}{*}{05/23} & \multirow{3}{*}{05/25} & Fuji        & 14 & 70 \\
                      &                        &                        &                        & Gala        & 14 & 70 \\
                      &                        &                        &                        & Honeycrisp  & 14 & 70 \\
\addlinespace[0.2cm]
\multirow{3}{*}{2022} & \multirow{3}{*}{05/22} & \multirow{3}{*}{05/24} & \multirow{3}{*}{05/26} & Fuji        & 0  & 70 \\
                      &                        &                        &                        & Gala        & 0  & 70 \\
                      &                        &                        &                        & Honeycrisp  & 0  & 70 \\
\addlinespace[0.2cm]
\multirow{1}{*}{2023} & \multirow{1}{*}{05/19} & \multirow{1}{*}{05/21} & \raisebox{0.3ex}{\shortstack{Freeze\\Occurred}} & Honeycrisp  & 70 & 0 \\
\hline
\addlinespace[0.5ex]
\multicolumn{4}{@{}c@{}}{} & Total & 112 & 420 \\
\hline \hline
\end{tabularx}
\label{table:dataset_div}
\vspace{-10pt}
\end{table*}

\subsubsection{Train and Evaluation Splits}
All imaged clusters were split into training and evaluation clusters, as shown in Table~\ref{table:dataset_div}. The training clusters were used to train and validate the Mask-RCNN network as well as tune the parameters for fruitlet clustering and temporal association. As discussed in Sec.~\ref{sec:cluster_sample}, because we use 70 clusters to predict thinning response, 14 of the 84 clusters per varietal imaged in 2021 were used for training, and the rest were used for evaluation. Additionally, a freeze occurred during our data collection in 2023, ending the growth of the fruitlets. As a result, clusters imaged in 2023 were used for training. All clusters imaged in 2022 were used for evaluation, which we use to demonstrate our method's ability to generalize to datasets taken in different years without the need to fine-tune any of the models. The final evaluation data includes 6 sets of 70 clusters evenly distributed over 2021 and 2022 Fuji, Gala, and Honeycrisp varietals.

\subsubsection{Annotation Labeling}\label{sec:annotation_labelling}
To train the Mask-RCNN network, 600 images were randomly selected from the images of the training clusters. Bounding boxes were labeled around every fruitlet to train the Mask-RCNN bounding box classification and regression heads, as discussed in Sec.~\ref{sec:seg_disp}. From the 600 images, 100 were randomly selected and used to manually hand-segment 1000 fruitlets to train the mask head. All data was then divided into training, validation, and test sets to train the network, with a 70/15/15 split respectively. A total of 600 images were used to train the bounding box heads because manual annotation of dense orchard scenes is highly time-consuming, particularly for apple fruitlets. This number is comparable with other datasets used in similar fruitlet detection studies (\cite{Checola, KHANAL20238914}). Additionally, our illumination-invariant camera system (\cite{9636542}) provides more consistent image quality under varying lighting conditions, which reduces the need for large training datasets compared to conventional camera systems. As discussed in Sec.~\ref{sec:seg_disp}, using 1000 manually segmented fruitlets across 100 images for mask training enables a two-stage training process that significantly reduces the need for exhaustive pixel-level annotation. 

\begin{figure*}
\centering
\includegraphics[width=0.99\linewidth]{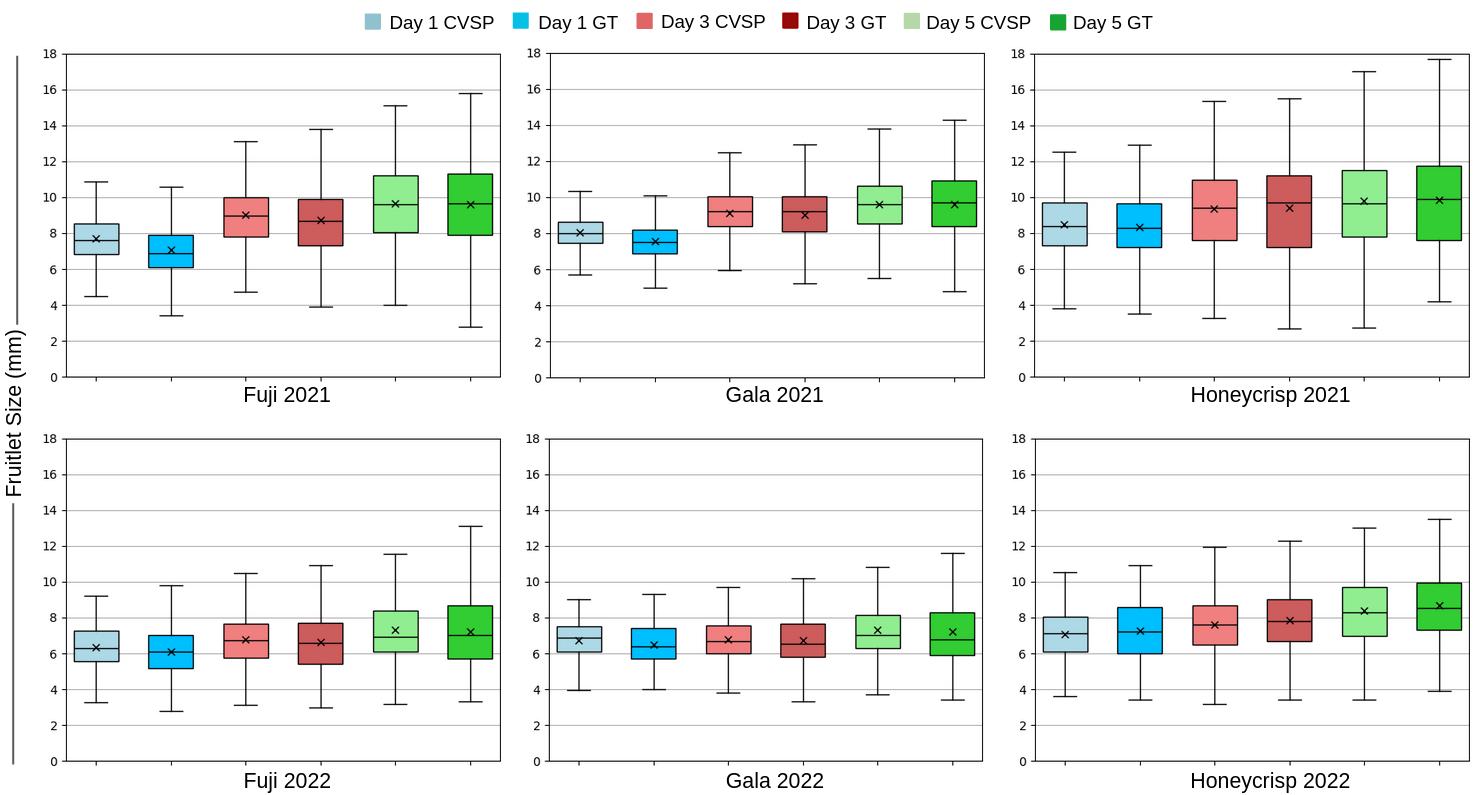}
\caption{Distribution of measured sizes for our computer vision sizing pipeline (CVSP) and ground-truth (GT). The mean and median sizes are indicated by the x and horizontal bar respectively.}
\label{fig:size_dist_comp}
\vspace{-15pt}
\end{figure*}
\section{Results}\label{sec:results}
\subsection{Caliper Measurement Variation}
Taking measurements with calipers naturally results in random errors. This is because the measurement will vary as the caliper is rotated around the fruitlet's asymmetrical shape. As well, the measurements depend on how hard the caliper is closed around the fruitlet, which can easily change from day to day and from person to person. To quantitatively assess 
variation, 45 fruitlets equally distributed between small ($<$~8mm), medium ($>$~8mm and $<$~12mm) and large ($>$~12mm) sizes were collected. For each fruitlet, we measured its diameter 5 times using calipers, and recorded the difference between the maximum and minimum sizes. The average differences for the small, medium, and large fruitlets were 0.50mm, 0.34mm, and 0.71mm respectively. The maximum differences were 2.31mm, 0.52mm, and 1.85mm respectively. All of these measurements were performed by the same person to isolate inconsistencies from single user variation. We expect greater variability if multiple individuals performed these measurements. These inconsistencies demonstrate the need for alternative solutions.

\subsection{Fruitlet Sizing}\label{subsec:sizing_results}
We evaluate our fruitlet sizing approach, removing outliers from the computer vision sizing pipeline that fall more than three standard deviations outside the mean. The number of outliers removed is less than 0.25\% of all fruitlets sized. The distribution of measured sizes using our computer vision sizing pipeline (CVSP) and the ground-truth caliper measurements (GT) are reported in Fig. \ref{fig:size_dist_comp}. While the computer vision pipeline produces slightly larger results on Day 1 for most datasets, the size and growth trends are similar across the five day period with regards to mean, median, and size distribution.

\begin{table*}
\scriptsize
\centering
\caption{Sizing results. For simplicity, the highest and second highest scoring All day results are highlighted in \hl{yellow} and \hlblue{blue} respectively.}
\newcolumntype{C}{>{\centering\arraybackslash}X}
\begin{tabularx}{\linewidth}{@{}lCCCCCCCCCCCC@{}}
\hline \hline
\addlinespace[0.6ex]
\multicolumn{13}{c}{\textbf{2021}} \\
\midrule
 & \multicolumn{4}{c}{Fuji} & \multicolumn{4}{c}{Gala} & \multicolumn{4}{c}{Honeycrisp} \\
Day & 1 & 3 & 5 & All & 1 & 3 & 5 & All & 1 & 3 & 5 & All \\
\midrule
\multicolumn{13}{l}{\textbf{CVSP (ours)}} \\
MAE (mm)      & 0.752 & 0.587 & 0.632 & \hlblue{0.656} & 0.661 & 0.612 & 0.661 & \hlblue{0.650} & 0.607 & 0.693 & 0.726 & \hlblue{0.675} \\
MAPE (\%)     & 11.8  & 7.46  & 7.91  & \hlblue{9.06}  & 9.25   & 7.45   & 7.30   & \hlblue{8.07}   & 8.48   & 8.75   & 8.72  & \hlblue{8.66} \\
Recall (\%)   & 88.2  & 93.6  & 91.4  & \hlblue{91.2}  & 97.1  & 98.6  & 97.4  & \hlblue{97.3} & 96.0  & 98.1  & 98.1  & \hlblue{97.0} \\
$R^2$ Score   & 0.841 & 0.889 & 0.881 & \hl{0.892} & 0.570 & 0.735 & 0.743 & \hlblue{0.745} & 0.850 & 0.855 & 0.876 & \hlblue{0.866} \\
\midrule
\multicolumn{13}{l}{\textbf{SI-CVSP}} \\
MAE (mm)      & 0.737 & 0.606  & 0.620  & \hl{0.653} & 0.658 & 0.667 & 0.725 & 0.685 & 0.629 & 0.714 & 0.826 & 0.722 \\
MAPE (\%)     & 11.5  & 7.64  & 7.52  & \hl{8.88} & 9.20  & 7.99 & 7.83 & 8.37 & 8.65 & 8.79 & 9.37 & 9.86 \\
Recall (\%)   & 86.1  & 92.9  & 90.0  & 89.8 & 95.3  & 97.7 & 95.6 & 95.7 & 95.1 & 96.9 & 96.4 & 95.7 \\
$R^2$ Score   & 0.836 & 0.858 & 0.878 & \hlblue{0.882} & 0.555 & 0.663 & 0.763 & 0.656 & 0.824 & 0.840 & 0.823 & 0.832 \\
\midrule
\multicolumn{13}{l}
{\textbf{FS-CVSP}} \\
MAE (mm)      & 0.776 & 0.590  & 0.641  & 0.689 & 0.654 & 0.600 & 0.655 & \hl{0.645} & 0.602 & 0.677 & 0.735 & \hl{0.672} \\
MAPE (\%)     & 12.1  & 7.49  & 8.05  & 9.61 & 9.16  & 7.28 & 7.23 & \hl{8.00} & 8.43 & 8.48 & 8.82 & \hl{8.60 }\\
Recall (\%)   & 88.1  & 94.0  & 91.8  & \hl{91.8} & 97.1  & 98.0 & 97.4 & \hlblue{97.3} & 96.3 & 98.1 & 98.1 & \hl{97.2} \\
$R^2$ Score   & 0.832 & 0.888 & 0.867 & 0.853 & 0.581 & 0.757 & 0.745 & \hl{0.748} & 0.852 & 0.871 & 0.874 & \hl{0.872} \\
\midrule
\multicolumn{13}{l}{\textbf{SGBM-CVSP}} \\
MAE (mm)      & 0.709 & 1.07  & 1.19  & 0.991 & 0.963 & 1.18 & 1.49 & 1.21 & 1.02 & 1.36 & 1.39 & 1.26 \\
MAPE (\%)     & 10.7  & 12.3  & 12.8  & 12.0 & 12.8  & 13.1 & 15.1 & 13.6 & 12.3 & 14.5 & 13.4 & 13.5 \\
Recall (\%)   & 89.5  & 93.2  & 90.0  & 90.5 & 97.7  & 98.3 & 97.4 & \hl{97.7} & 95.7 & 97.8 & 97.1 & 96.4 \\
$R^2$ Score   & 0.700 & 0.624 & 0.710 & 0.703 & 0.243 & 0.402 & 0.400 & 0.421 & 0.611 & 0.644 & 0.680 & 0.649 \\
\midrule
\midrule
\multicolumn{13}{c}{\textbf{2022}} \\
\midrule
 & \multicolumn{4}{c}{Fuji} & \multicolumn{4}{c}{Gala} & \multicolumn{4}{c}{Honeycrisp} \\
Day & 1 & 3 & 5 & All & 1 & 3 & 5 & All & 1 & 3 & 5 & All \\
\midrule
\multicolumn{13}{l}{\textbf{CVSP (ours)}} \\
MAE (mm)      & 0.630 & 0.598 & 0.626 & \hl{0.617} & 0.655 & 0.627 & 0.589 & \hl{0.621} & 0.705 & 0.756 & 0.795 & \hl{0.762} \\
MAPE (\%)     & 11.9 & 10.3 & 9.80 & \hl{10.7} & 10.7 & 9.67   & 8.71  & \hl{9.68}  & 10.2  & 10.5 & 9.88   & \hl{10.4} \\
Recall (\%)   & 88.4  & 91.4  & 91.5  & \hlblue{89.9}  & 92.1  & 94.9  & 96.7  & \hlblue{94.1}  & 96.3  & 95.9  & 96.6  & \hl{96.5} \\
$R^2$ Score   & 0.703 & 0.752 & 0.824 & \hl{0.772} & 0.591 & 0.671 & 0.838 & \hl{0.719} & 0.604 & 0.675 & 0.764 & \hlblue{0.712} \\
\midrule
\multicolumn{13}{l}{\textbf{SI-CVSP}} \\
MAE (mm)      & 0.660 & 0.597 & 0.636 & \hlblue{0.629} & 0.666  & 0.680  & 0.628  & \hlblue{0.656}  & 0.747 & 0.813 & 0.932 & 0.835 \\
MAPE (\%)     & 12.3  & 10.3  & 9.91  & \hlblue{10.9}  & 10.7  & 10.1  & 8.87  & \hlblue{9.91}  & 10.7  & 10.9  & 11.1  & 11.0 \\
Recall (\%)   & 86.5  & 89.5  & 89.9  & 88.1  & 89.2  & 92.1  & 93.3  & 90.9  & 93.6  & 92.1  & 93.5  & 93.0 \\
$R^2$ Score   & 0.663 & 0.752 & 0.815 & 0.757 & 0.552 & 0.588 & 0.787 & 0.653 & 0.564 & 0.638 & 0.648 & 0.643 \\
\midrule
\multicolumn{13}{l}
{\textbf{FS-CVSP}} \\
MAE (mm)      & 0.630 & 0.635 & 0.644 & 0.635 & 0.673  & 0.724  & 0.602  & 0.667  & 0.711 & 0.789 & 0.787 & \hlblue{0.772} \\
MAPE (\%)     & 11.9  & 10.7  & 10.1  & \hlblue{10.9}  & 10.9  & 11.3  & 8.92  & 10.4  & 10.3  & 10.6  & 9.7  & \hlblue{10.5} \\
Recall (\%)   & 88.3  & 91.1  & 91.8  & \hl{90.0}  & 92.1  & 95.3  & 96.7  & \hl{94.6}  & 96.3  & 96.9  & 95.5  & \hl{96.5} \\
$R^2$ Score   & 0.704 & 0.731 & 0.809 & \hlblue{0.762} & 0.593 & 0.560 & 0.833 & \hlblue{0.691} & 0.615 & 0.673 & 0.775 & \hl{0.719} \\
\midrule
\multicolumn{13}{l}
{\textbf{SGBM-CVSP}} \\
MAE (mm)      & 0.768 & 0.816 & 0.866 & 0.817 & 1.18  & 1.20  & 1.03  & 1.12  & 0.898 & 0.872 & 0.925 & 0.904 \\
MAPE (\%)     & 13.4  & 13.4  & 12.8  & 13.3  & 18.1  & 18.5  & 14.7  & 16.9  & 12.6  & 11.4  & 11.7  & 11.9 \\
Recall (\%)   & 68.1  & 56.2  & 69.2  & 64.5  & 58.3  & 69.7  & 76.4  & 67.8  & 70.6  & 70.3  & 73.2  & 71.6 \\
$R^2$ Score   & 0.576 & 0.636 & 0.703 & 0.665 & 0.335 & 0.405 & 0.612 & 0.474 & 0.535 & 0.611 & 0.697 & 0.666 \\
\hline \hline
\end{tabularx}
\label{table:results}
\vspace{-5pt}
\end{table*}

\begin{figure*}
\centering
\includegraphics[width=0.99\linewidth]{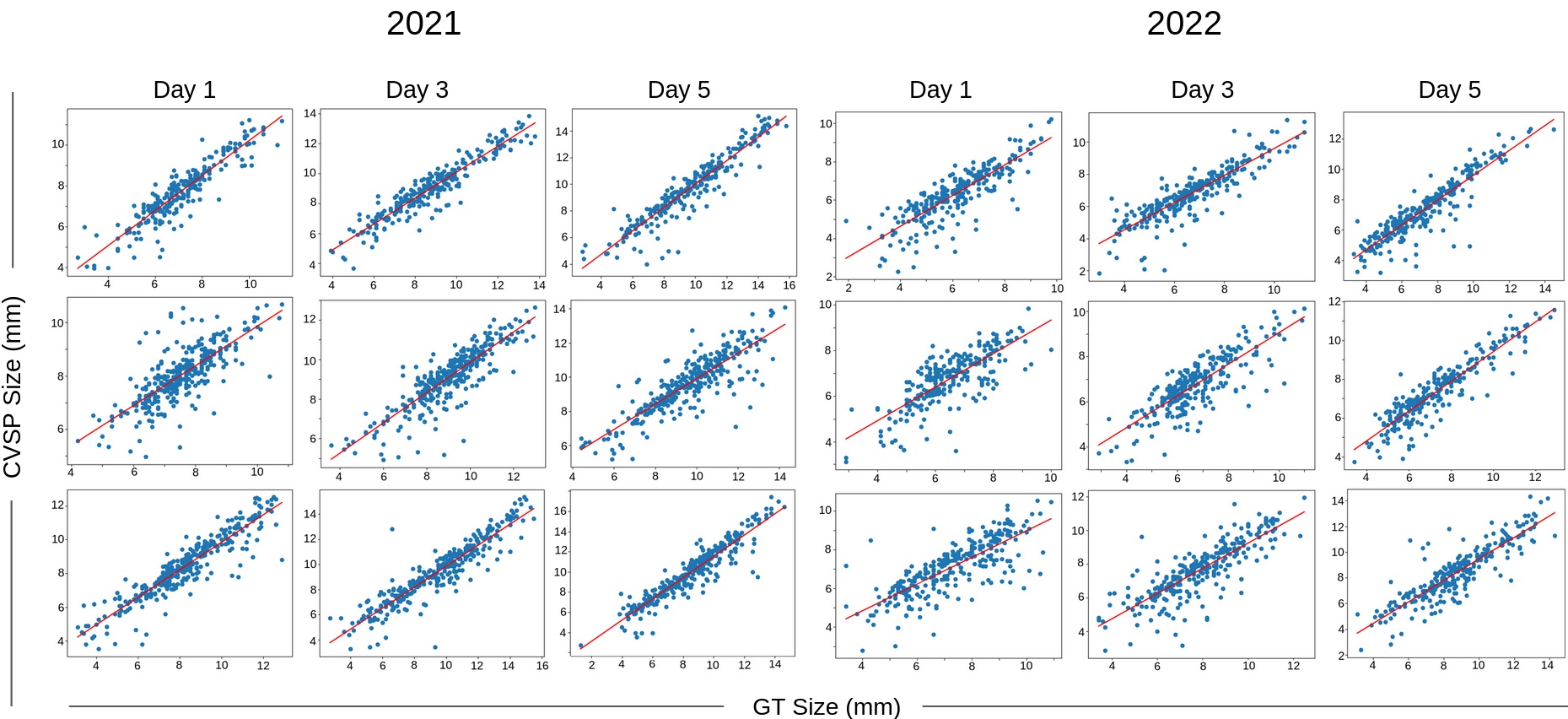}
\caption{Linear fit plots between our CVSP sizes and ground-truth sizes. Top, middle, and bottom rows are Fuji, Gala, and Honeycrisp respectively.}
\label{fig:r2_comp}
\vspace{-15pt}
\end{figure*}

\begin{figure*}
\centering
\includegraphics[width=0.99\linewidth]{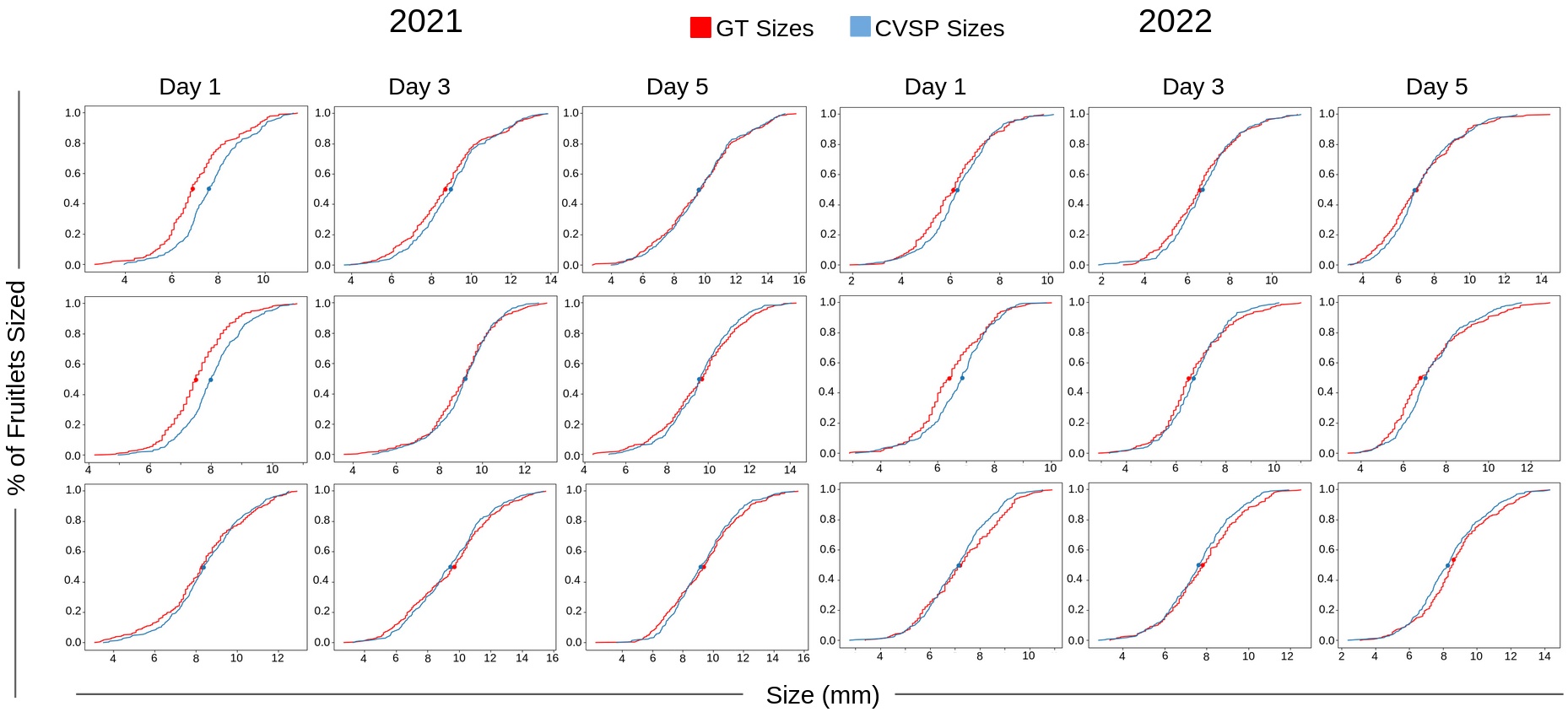}
\caption{Cumulative distribution functions of sizes. The blue and red dots indicate the median values of the CVSP and GT sizes respectively. Top, middle, and bottom rows are Fuji, Gala, and Honeycrisp respectively.}
\label{fig:cdf_comp}
\vspace{-15pt}
\end{figure*}

Table \ref{table:results} shows the Mean Absolute Error (MAE), the Mean Absolute Percentage Error (MAPE), the recall (the percentage of fruitlets sized), and the $R^2$ linear fit score of the computer vision sizes compared against the caliper method for each day, as well as across all days. All linear fit plots can be seen in Fig.~\ref{fig:r2_comp}. The MAE is consistently under 1mm, and the MAPE remains below 10\% for the 2021 varietals and around 10\% for the 2022 varietals. This difference is likely because no 2022 varietals were used to train the Mask-RCNN network, and results could be improved with a more diverse dataset. Additionally, we are able to size approximately 88\% - 99\% of fruitlets each day, and achieve $R^2$ scores between 0.570 and 0.899 with an interquartile range of 0.707 and 0.892.

As expected, fruitlets sized on the first day typically have a higher MAPE and lower $R^2$ score compared to the later days. One possible reason for the higher MAPE and lower $R^2$ scores on the first day is because it is more difficult to accurately segment smaller fruitlets as their flowery top covers a greater percentage of the fruit, resulting in fewer valid disparity values and more inconsistent ellipse shapes. On the other hand, the errors may stem from the inconsistencies in measuring ground-truth. The size variations from using calipers would have a more significant effect when the fruitlets are small, as differences in fruitlet turgor and firmness at early growth stages can alter the measured diameter depending on how tightly the caliper is applied.

We also plot the cumulative distribution functions (CDF) of fruitlet sizes in Fig.~\ref{fig:cdf_comp}. The CDFs follow similar shapes across all varietals and days. The most noticeable difference is on Day 1 for most varietals, where there is a horizontal shift between the CVSP and ground-truth distributions. This is again because of the difficulty in sizing smaller fruit.

We investigate the effects of using only a single image for every cluster (SI-CVSP), replacing RAFT-Stereo with a state-of-the-art disparity estimation method FoundationStereo presented by \cite{wen2025foundationstereozeroshotstereomatching} (FS-CVSP), and replacing RAFT-Stereo with the classical disparity estimation method SGBM (\cite{SGBM_2008}) (SGBM-CVSP). As shown in Table~\ref{table:results}, using a second image generally improves performance over using a single image by increasing the number of visible fruitlets in cases of occlusion, and consistently yields higher recall. RAFT-Stereo and FoundationStereo produce comparable results. While we opt to use RAFT-Stereo because it is slightly faster (Table~\ref{table:size_times}) and less computationally expensive, this demonstrates that the choice is flexible and future state-of-the-art deep-learning-based disparity estimation networks could be integrated. SGBM results in higher MAE and MAPE values and lower $R^2$ scores. This is likely because SGBM is unable to assign a disparity value to every pixel. As well, the performance of its disparity estimation is sensitive to the algorithm parameters used. Another interesting behavior is the SGBM pipeline sizes a significantly fewer number of fruitlets for the 2022 varietals. This is likely due to the combination of the segmentation network not being trained on 2022 data and weaker disparity estimations. 

\begin{table*}[!t]
\scriptsize
\centering 
\caption{Evaluation of different clustering methods. \hl{Highlighted} values indicate best performing metric per varietal.}
\begin{tabularx}{\textwidth}{@{}l l *{6}{>{\centering\arraybackslash}X}@{}}
\hline\hline
& & \multicolumn{3}{c}{2021} & \multicolumn{3}{c}{2022} \\
& & Fuji & Gala & Honeycrisp & Fuji & Gala & Honeycrisp \\
\hline
\multirow{3}{*}{Ours}
& Precision (\%) & 97.6 & 99.6 & 98.9 & 97.2 & 98.2 & 99.1 \\
& Recall (\%)    & \hlyellow{86.4} & \hlyellow{95.8} & \hlyellow{95.6} & 87.2 & \hlyellow{92.6} & \hlyellow{94.2} \\
& F1 (\%)        & \hlyellow{91.7} & \hlyellow{97.7} & \hlyellow{97.3} & \hlyellow{91.9} & \hlyellow{95.3} & \hlyellow{96.6} \\
\hline
\multirow{3}{*}{LCD}
& Precision (\%) & \hlyellow{99.0} & \hlyellow{99.7} & \hlyellow{99.7} & \hlyellow{99.4} & \hlyellow{99.4} & \hlyellow{99.2} \\
& Recall (\%)    & 77.7 & 90.1 & 94.8 & 78.7 & 86.0 & 84.5 \\
& F1 (\%)        & 87.1 & 94.7 & 97.2 & 87.8 & 92.2 & 91.3 \\
\hline
\multirow{3}{*}{DBSCAN}
& Precision (\%) & 97.2 & 98.5 & 96.9 & 96.8 & 96.1 & 98.0 \\
& Recall (\%)    & 83.4 & 95.0 & \hlyellow{95.6} & \hlyellow{87.3} & 92.5 & 93.8 \\
& F1 (\%)        & 89.8 & 96.7 & 96.2 & 91.8 & 94.3 & 95.9 \\
\hline
\multirow{3}{*}{HDBSCAN}
& Precision (\%) & 83.4 & 90.5 & 85.8 & 89.3 & 91.7 & 92.0 \\
& Recall (\%)    & 76.5 & 93.2 & 88.9 & 67.7 & 73.5 & 60.3 \\
& F1 (\%)        & 79.8 & 91.8 & 87.3 & 77.0 & 81.6 & 72.9 \\
\hline\hline
\end{tabularx}
\label{table:cluster_pred}
\end{table*}

\begin{table*}
\scriptsize
\centering
\caption{Evaluation of different association methods. 
\hlyellow{Highlighted} values indicate the best performing metric per varietal.}
\begin{tabularx}{\linewidth}{@{}QJBDEFGH@{}} 
\hline\hline 
\addlinespace[0.1cm]
\multirow{2}{*}{{\shortstack[c]{\vspace{11pt}\\ \ }}} & &
& 2021 & &
& 2022 & \\
\addlinespace[0.1cm]
& & Fuji & Gala & Honeycrisp &
Fuji & Gala & Honeycrisp\\[0.5ex]
\hline
\addlinespace[0.1cm]
\multirow{2}{*}{{\shortstack[c]{\vspace{10pt}\\ Ours }}}
& Precision~(\%) & \hlyellow{94.7} & \hlyellow{95.3} & \hlyellow{95.0} & \hlyellow{90.2} & \hlyellow{88.7} & \hlyellow{92.3}\\
\addlinespace[0.1cm]
& Recall~(\%) & 86.8 & \hlyellow{92.2} & 91.1 & 84.0 & \hlyellow{80.0} & 86.6\\
\addlinespace[0.1cm]
& F1~(\%) & \hlyellow{90.6} & \hlyellow{93.8} & \hlyellow{93.0} & 87.0 & \hlyellow{84.1} & \hlyellow{89.3}\\
\addlinespace[0.1cm]
\hline
\addlinespace[0.1cm]
\multirow{2}{*}{{\shortstack[c]{\vspace{10pt}\\ Desc-Assoc}}} & Precision~(\%) & 90.1 & 91.5 & 94.0 & 87.7 & 81.4 & 87.6\\
\addlinespace[0.1cm]
& Recall~(\%) & \hlyellow{87.2} & 91.7 & \hlyellow{91.7} & \hlyellow{87.9} & 78.6 & \hlyellow{86.7}\\
\addlinespace[0.1cm]
& F1~(\%) & 89.0 & 91.6 & 92.8 & \hlyellow{87.8} & 80.0 & 87.1\\
\addlinespace[0.1cm]
\hline
\addlinespace[0.1cm]
\multirow{2}{*}{{\shortstack[c]{\vspace{10pt}\\ Size-Assoc}}} & Precision~(\%) & 51.9 & 36.6 & 48.9 & 48.9 & 35.1 & 42.3\\
\addlinespace[0.1cm]
& Recall~(\%) & 55.3 & 37.8 & 50.3 & 51.9 & 37.6 & 43.9\\
\addlinespace[0.1cm]
& F1~(\%) & 53.6 & 37.2 & 49.6 & 50.4 & 36.3 & 43.1\\
\addlinespace[0.1cm]
\hline
\end{tabularx}
\label{table:assoc_results}
\vspace{-5pt}
\end{table*}

\subsection{Clustering}\label{sec:clustering_performance}
We evaluate the performance of our clustering method presented in Sec.~\ref{sec:fruitlet_cluster} and compare it against unmodified Louvain Community Detection (LCD)~\cite{community_cluster} in addition to DBSCAN~\cite{dbscan} and HDBSCAN~\cite{hdbscan}. The results can be seen in Table~\ref{table:cluster_pred}. Our multi-stage Louvain Community Detection approach achieves the highest recall in five out of six datasets and the highest F1 score across all varietals. Unmodified Louvain Community Detection achieved the highest precision across all datasets, but had lower recall scores due to its tendency to separate clusters. Across most varietals, DBSCAN demonstrated competitive performance compared to our multi-stage approach. This suggests it could be substituted into our pipeline with only a slight reduction in clustering performance, if an off-the-shelf method is preferred.

\subsection{Temporal Association}\label{sec:temp_reg_res}
We assess the effectiveness of our temporal association method discussed in Sec.~\ref{sec:cross_day_association} and evaluate it against two baselines. The first baseline is a naive approach that associates fruitlets based solely on the order of size (Size-Assoc). This association method is often used in practice when growers use the modified Ferri version of the Fruitlet Growth Model~\footnote[10]{\url{https://ag.umass.edu/sites/ag.umass.edu/files/fruit/predictingfruitsetmodelferri.pdf}} and apps such as Malusim~\footnote[11]{\url{https://malusim.org/}} in order to save time from having to manually associate fruitlets.

The second baseline is the method presented by Riccardi \textit{et al.}~\cite{riccardi} (Desc-Assoc)\footnote[12]{Because no open source code was available, we re-implemented the association approach of~\cite{riccardi} based on the available details}. As described in the original paper, the weighting parameters are optimized using Optuna (\cite{optuna_2019}) on the training clusters presented in Table~\ref{table:dataset_div}. Three adjustments were made to the original approach. First, the radius term is removed since the radii between detected fruitlets at different time steps can vary significantly due to growth and partial occlusions. Second, initial point cloud alignment is performed using only G-ICP~\cite{g-icp} and not RANSAC~\cite{ransac}. This is because it is challenging to consistently detect matching features in pairs of images. However, G-ICP is run only on a local volume surrounding the detected target cluster, rather than the entire scene, which improved the registration results. Lastly, all neighboring fruitlets in the cluster contribute to the descriptor, as opposed to using a fixed number of nearest neighbors of all fruitlets in the scene. The reason for this is the number of fruitlets in a cluster often varies from three to six, and using all detected fruitlets in the scene results in poor performance due to variations in camera pose and missed detections.

The results can be seen in Table~\ref{table:assoc_results}. Our method achieves the highest precision across all varietals and years. Additionally, our approach has the highest F1 scores in five out of six datasets, with values between 84.1\% and 93.8\%. Unsurprisingly, the naive method of associating fruitlets based on size performed poorly. While expected, it is worth quantifying to understand the limitations of a current method used in practice.

\begin{table*}[!t]
{\fontsize{7}{9}\selectfont  
\setlength{\tabcolsep}{2pt}  
\centering
\caption{Thinner Response Prediction and Association Accuracy Results}
\begin{tabularx}{\linewidth}{@{}lCCCCCCC@{}}
\hline\hline
 & \multicolumn{3}{c}{\textbf{2021}} & \multicolumn{3}{c}{\textbf{2022}} & \textbf{Avg} \\
 & Fuji & Gala & Honeycrisp & Fuji & Gala & Honeycrisp & \\
\hline
GT Persist/Abscise (\%) 
& 53.0/47.0 & 48.0/52.0 & 27.2/72.8 
& 37.2/62.8 & 30.5/69.5 & 42.4/57.6 
& -- \\
\hline
\textbf{FGMP (ours)} &&&&&&& \\
Persist/Abscise (\%) 
& 52.6/47.4 & 41.7/58.3 & 26.7/73.3 
& 37.5/62.5 & 24.5/74.5 & 35.0/65.0 
& -- \\
(APD) (\%) 
& 0.4 & 6.3 & 0.5 
& 0.3 & 6.0 & 7.4 
& 3.48 \\
E2E Recall (\%) 
& 82.7 & 90.9 & 89.0 
& 75.4 & 76.1 & 84.0 
& 83.0 \\
\hline

\textbf{SI-FGMP} &&&&&&& \\
Persist/Abscise (\%) 
& 51.9/48.1 & 40.8/59.2 & 23.7/76.3 
& 38.8/61.2 & 19.8/80.2 & 30.4/69.6 
& -- \\
(APD) (\%) 
& 1.1 & 7.2 & 3.5 
& 1.6 & 10.7 & 12 
& 6.02 \\
E2E Recall (\%) 
& 81.4 & 90.9 & 89.0
& 76.6 & 75.5 & 84.0 
& 82.9 \\

\hline
\textbf{KA-FGMP} &&&&&&& \\
Persist/Abscise (\%) 
& 49.6/50.4  & 41.2/58.8 & 26.3/73.7 
& 36.0/64.0 & 23.5/76.5 & 33.0/67.0 
& -- \\
(APD) (\%) 
& 3.4 & 6.8 & 0.9 
& 1.2 & 7.0 & 9.4 
& 4.78 \\
\hline\hline
\end{tabularx}
\label{table:e2e_results}
}
\end{table*}

\begin{table*}
\scriptsize
\centering
\caption{Speed Evaluation}
\begin{tabularx}{\linewidth}{@{}AJJJ@{}} 
\hline\hline 
\addlinespace[0.1cm]
\ & Mean (s) & Median (s) & Std (s) \\
\hline
\addlinespace[0.1cm]
FoundationStereo & 2.68 & 2.68 & 0.0117 \\
\hline
\addlinespace[0.1cm]
RAFT-Stereo & 2.38 & 2.40 & 0.0316 \\
\addlinespace[0.1cm]
Mask-RCNN & 0.126 & 0.121 & 0.0293 \\
\addlinespace[0.1cm]
Sizing & 0.00437 & 0.00416 & 0.00160 \\
\addlinespace[0.1cm]
Fruitlet Clustering & 0.188 & 0.186 & 0.0136 \\
\addlinespace[0.1cm]
Association & 0.362 & 0.355 & 0.0367 \\
\addlinespace[0.1cm]
Total & 3.06 & 3.07 & 0.0367\\
\hline\hline
\end{tabularx}
\label{table:size_times}
\vspace{-5pt}
\end{table*}

\subsection{Fruitlet Association and Thinner Response Prediction}
We evaluate the end-to-end performance of our Fruitlet Growth Measurement Pipeline (FGMP) and its ability to estimate persist and abscission rates. Additionally, we compare its performance to results obtained using only a single image for every cluster (SI-FGMP) and using known associations (KA-FGMP). The former provides insight into the extent to which a second image improves accuracy, and the latter into how well our system performs with no association error, which allows us to demonstrate the performance where the grower chooses to manually associate the fruitlets.

The full results can be seen in Table~\ref{table:e2e_results}. We report the end-to-end pipeline's ability to predict thinning response (Persist / Abscise) as well as the absolute percentage difference (APD) compared to the ground-truth values. Additionally, we report the end-to-end recall (E2E Recall) for the FGMP pipeline. This is different from the recall presented in Table~\ref{table:assoc_results}, as the end-to-end recall also accounts for false negatives resulting from incorrect clustering. The precision values are the same as in Table~\ref{table:assoc_results}.

Our full end-to-end pipeline is able to achieve an average APD of 3.48\%.
This result best demonstrates the effectiveness of our approach. Apple growers would be able to draw similar conclusions about when to spray using our fully automated pipeline, which requires no manual sizing or fruitlet identification. As expected, using only a single image per cluster resulted in a higher APD across all datasets, likely due to increased occlusions that make fruitlets more difficult to accurately size. Interestingly, using known associations results in a slightly higher average APD of 4.78. Possible explanations for this behavior include cumulative errors throughout the pipeline or inaccurate ground-truth measurements. This further demonstrates the contributions of a full end-to-end pipeline, as the benefit of reduced labor will have a greater impact than manually associating individual fruitlets. Regardless of which method is chosen, both result in similar spraying decision making strategies compared to using hand-calipers.

\subsection{Speed Evaluation}
We assess the speed of our sizing and association pipelines. Table~\ref{table:size_times} summarizes the distribution of sizing times from image input to sizing output. The evaluation was performed on an NVIDIA GeForce RTX 3090 GPU. The average processing time is approximately 3.06s, with RAFT-Stereo accounting for the majority of the processing time with an average of 2.38s. We also report the speed of integrating FoundationStereo for comparison. The speed of the full pipeline will be affected by the GPU used.

We asked apple growers how long it usually takes to size fruitlets with calipers. They reported that it takes approximately 30s per cluster, which we confirmed when collecting ground-truth measurements. Based on this preliminary study, our computer vision pipeline sized fruitlets approximately ten times faster than hand-caliper measurements.

Because the field data used in this study was collected from longer video sequences to allow for the manual selection of one to two representative images per cluster (as discussed in Sec.~\ref{subsec:data_collection}), and because some data was captured using a mobile robotic platform (Sec.~\ref{sec:robot_exp}), we do not report a full end-to-end timing comparison that includes walking between clusters to record growth rates versus capturing images. However, we assume that the additional time required to walk between clusters and capture one AprilTag image and one to two cluster images would be negligible compared to the overall time spent walking itself.

\begin{table*}[!t]
\scriptsize
\centering
\caption{RCVSP Sizing Results}
\begin{tabularx}{\linewidth}{@{}AJJJJ@{}} 
\hline\hline 
\addlinespace[0.1cm]
\ & Day 1 & Day 3 & Day 5 & All \\
\hline
\addlinespace[0.1cm]
MAE (mm) & 0.834 & 0.911 & 0.981 & 0.919 \\
\addlinespace[0.1cm]
MAPE (\%) & 14.2 & 13.6 & 13.2 & 13.9 \\
\addlinespace[0.1cm]
Recall~(\%) & 91.5 & 90.1 & 81.6 & 88.4 \\
\addlinespace[0.1cm]
\hline\hline
\end{tabularx}
\label{table:robot_res}
\end{table*}

\begin{figure}
\centering
\includegraphics[width=0.99\linewidth]{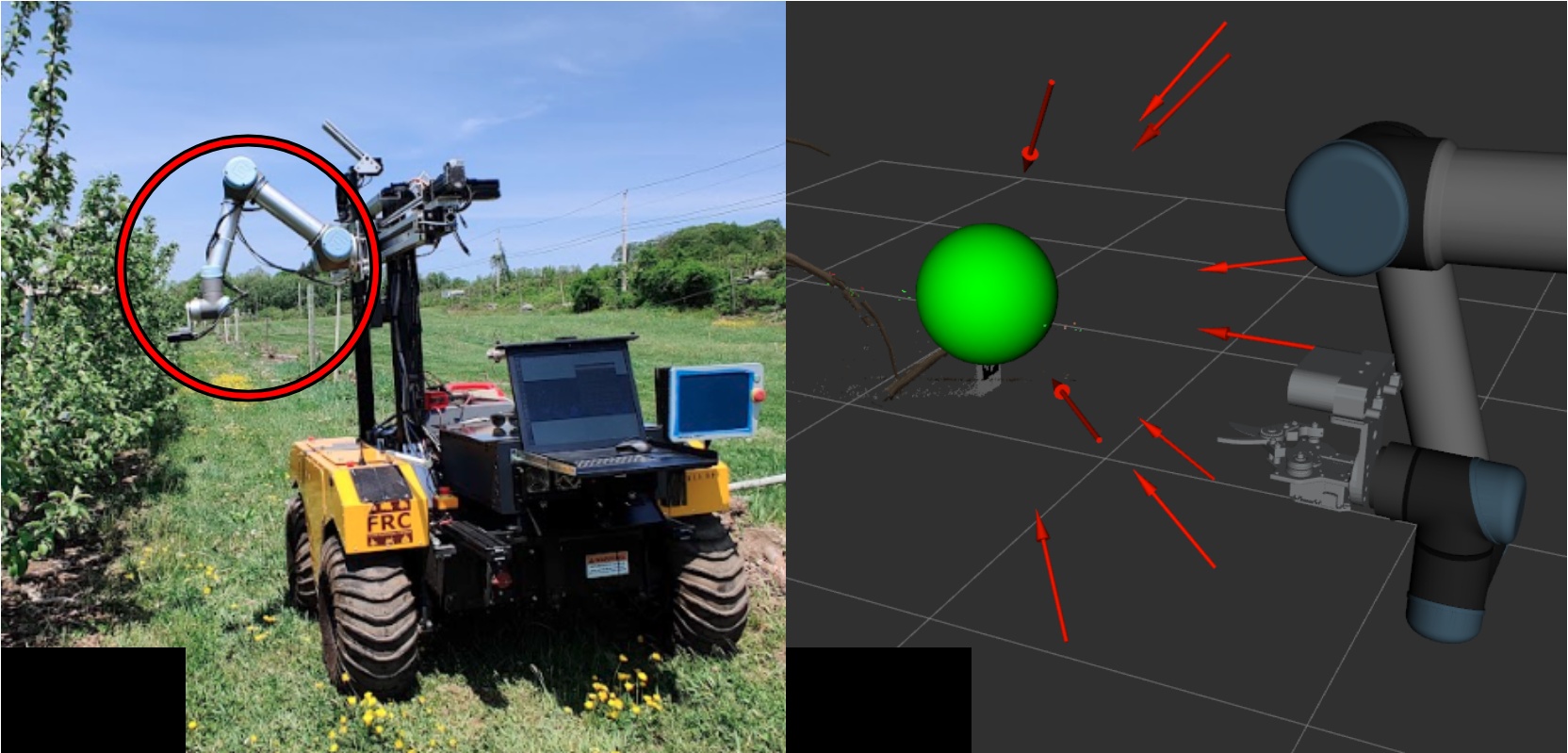}
\put(-227, 5){\textcolor{white}{\small{(a)}}}
\put(-110, 5){\textcolor{white}{\small{(b)}}}
\vspace{-5pt}
\caption{(a) Flash stereo camera attached to 7 DoF robotic arm. (b) Robot motion path where 15 images are collected at evenly distributed viewpoints on a hemisphere around the detected cluster.}
\label{fig:robot_path}
\vspace{-15pt}
\end{figure} 

\subsection{Robot Experiment}\label{sec:robot_exp}
We test our sizing pipeline on images collected by a robotic system in the field. The flash stereo camera is attached to a 7 DoF robotic arm consisting of a UR5 and linear slider (\cite{silwal2021bumblebee}). Images of 30 gala clusters were taken at the University of Massachusetts Amherst Cold Spring Orchard on 3 different days spanning a five-day window: 05/22/2022, 05/24/2022, and 05/26/2022, which we denote as Day 1, Day 3, and Day 5 respectively. The robot was programmed to follow a predefined hemispherical motion path around the detected AprilTag, collecting images at viewpoints equally distributed throughout the motion path~(Fig. \ref{fig:robot_path}). During image capture, the distance between the stereo camera and the centroid of the target cluster was maintained at approximately 40cm.

The same process described in Sec.~\ref{sec:methodology} was used to size the fruitlets. To associate detections in different images, we first cluster the fruitlets following the method presented in Sec.~\ref{sec:fruitlet_cluster}. However, because the AprilTag might not be visible in every image, we determine the position of the target cluster in the first image and use the robot kinematics to determine its location in subsequent images. The closest cluster in each subsequent image is determined to be the target cluster. We then use the same association method described in Sec.~\ref{sec:cross_day_association}, with the addition of using the robot kinematics to guide the initial transform of ICP. For each fruitlet, the two images with the greatest segmented area were used to determine its size.

\begin{figure}
\centering
\includegraphics[width=0.99\linewidth]{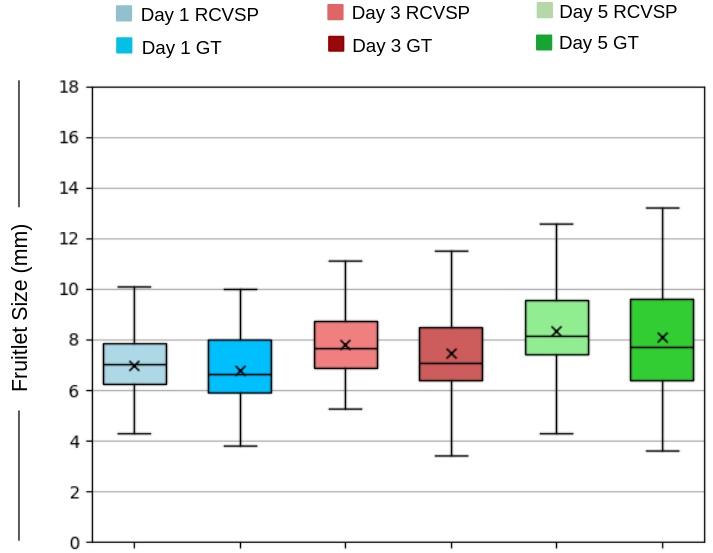}
\vspace{-5pt}
\caption{Distribution of measured sizes for our robot computer vision sizing pipeline (RCVSP) and the caliper method (GT). The mean and median sizes are indicated by the x and horizontal bar respectively.}
\label{fig:robot_size_dist}
\vspace{-15pt}
\end{figure}

We record the size distributions (Fig. \ref{fig:robot_size_dist}) for measurements taken by both our computer vision sizing pipeline with robot captured images (RCVSP) and the caliper method (GT). Across all days, the sizes produced by the computer vision pipeline are larger on average than those produced by the caliper method, and the distributions are often narrower. This is because the robot is naively following a pre-set path and not reasoning about where to capture an image. As a result, smaller fruitlets are likely to be fully occluded.

The MAE, MAPE, and recall (the percentage of fruitlets sized) are presented in Table \ref{table:robot_res}. While the MAE remains under 1mm, the MAPE is higher on average compared to the handheld camera results presented in Sec.~\ref{subsec:sizing_results}. This is because of the partial occlusion of the fruitlets in the images as a result of following the pre-set path. Additionally, the percentage of fruitlets sized decreases as time goes on. This is a result of the larger fruitlets fully occluding the smaller ones, which has a more significant effect as the fruitlets grow.
\vspace{-15pt}
\section{Discussion}\label{sec:discussion}
\subsection{Comparison to Related Works}

As discussed in Sec.~\ref{introduction}, several recent works have focused on developing computer vision systems to analyze apple fruitlets. However, these works focus on either detection or detection combined with sizing. Studies presented by \cite{WANG2021271, Qadri, KHANAL20238914, Sapkota} demonstrate the ability to detect and segment fruitlets using models such as YOLO and pix2pix, and size them using other photogrammetric or 3D reconstruction techniques. While these methods perform well for their intended tasks, none provide a complete end-to-end system for tracking fruitlet growth rates over time - a key requirement for applying the Fruitlet Growth Model to guide thinning decisions. These approaches analyze only single images rather than tracking fruitlets temporally across multiple days. It is worth noting that the detection and segmentation networks from these works could be integrated into our pipeline as modular alternatives to Mask-RCNN, offering flexibility to users who may prefer to trade reduced labeling effort for faster inference or different computational constraints.

The system most similar to ours is the one recently presented by \cite{Checola}, which sizes and counts clustered fruitlets directly from video sequences. Their approach clusters foreground and background fruitlets using hierarchical clustering and processes sequences of images with automated frame filtering and anomaly detection using Variational Autoencoders. Detections within sequences are matched using the Hungarian Algorithm, and fruitlets are sized using the dimensions of the predicted bounding boxes, the depth measurements from a RealSense camera, and the median size of the matched fruitlets within each sequence.

While video-based approaches have advantages, they also come with costs. Capturing and processing long sequences of images is substantially more time-consuming than capturing one or two static images per cluster, and increases the complexity of data collection in orchard environments. Although the Hungarian Algorithm is robust and simple to integrate, maintaining tracking accuracy over lengthy sequences in occluded environments can be challenging. By limiting the number of images used to one or two per cluster, we avoid these complications and reduce both the field collection time and computational processing requirements. This strategy, however, may slightly reduce sizing accuracy. Future studies could explore the optimal number of images per cluster to balance accuracy and efficiency.

We also compare our sizing results to those presented by \cite{Checola}. While \cite{Checola} report $R^2$ scores separately for apical, lateral, and spur Fuji fruitlets, the lateral fruitlets most closely match the size distribution of our dataset, ranging between 4–20mm. On their lateral fruitlet dataset, they report an $R^2$ score of 0.863, whereas our Fuji 2021 dataset achieves a slightly higher score of 0.892. Averaged across the Fuji, Gala, and Honeycrisp varietals, our 2021 datasets achieve $R^2$ scores of 0.892, 0.745, and 0.866 respectively, resulting in an overall average of 0.834. While these results are similar, the discrepancy may arise from their use of longer video sequences or from differences in varietals, environmental conditions, and data collection procedures. Our 2022 datasets exhibit slightly lower $R^2$ scores, likely because no 2022 data was included in the training split (Table~\ref{table:dataset_div}). In contrast, \cite{Checola} do not include a cross-season split, as all data was captured within the same time period and divided into 60/20/20 training, validation, and test sets. Further analysis over multiple growing seasons would be needed to fully assess how performance improves as additional data becomes available.

Most notably, \cite{Checola} do not address spatio-temporal association across days, which is essential for modeling fruitlet growth rates and predicting thinner response. The Pometa\footnote[8]{\url{https://igrownews.com/pometa-smart-crop-load-management-for-apple-growers/}} system, an industry-developed tool, also performs growth rate estimation; however, its methods are proprietary and not publicly available, and it relies on video-based image acquisition, which can take 24 hours to process (\cite{pometa_eval}). To the best of our knowledge, this study is among the first academic works to present an end-to-end computer vision system that performs fruitlet detection, sizing, clustering, and multi-day association for growth rate estimation and thinner response prediction. Furthermore, because our system is designed with modularity in mind, each component can be replaced with alternative methods as new state-of-the-art approaches are developed.

\subsection{Thinner Response Prediction}

Our end-to-end system achieves strong performance in predicting persist and abscission rates according to the Fruitlet Growth Model. Across six datasets spanning Fuji, Gala, and Honeycrisp varietals collected during the 2021 and 2022 growing seasons, our computer vision-based approach predicted persist and abscission rates with absolute percentage differences ranging from 0.4\% to 7.4\% compared to traditional hand-caliper measurements. The average absolute percentage difference across all datasets was 3.48\%, indicating that our automated system produces results highly consistent with the current manual standard.

These results demonstrate that our approach could provide an alternative to hand-caliper measurements for thinning response prediction. While additional validation across diverse orchards and growing conditions would strengthen confidence in widespread adoption, the close agreement between our predictions and those derived from manual measurements suggests that computer vision-based growth rate tracking has significant potential for practical deployment. With further evaluation across additional orchards, growing conditions, and varietals, computer vision systems could eventually replace manual caliper measurements, offering apple growers a faster and less labor-intensive method for making critical thinning decisions.

\subsection{Limitations}
While our approach produces promising results, there is still work needed in order to make it adoptable by growers. For one, our pipeline uses a unique and capable illumination-invariant camera system (\cite{9636542}). Although this could be replaced with a readily available commercial camera, such as the Intel\textsuperscript{\textcopyright} RealSense\textsuperscript{TM} used by \cite{Checola}, additional analysis would be required to evaluate the impact of varying illumination conditions on fruitlet appearance and overall measurement accuracy.

Another limitation of this study lies in the size and diversity of the training dataset. The Mask-RCNN network was trained on 600 images, which is comparable to datasets used in related fruitlet detection and sizing studies (\cite{Checola}, \cite{KHANAL20238914}). However, expanding the dataset could further improve segmentation performance, as suggested by the observed difference between 2021 and 2022 varietals—where the 2021 data, used for training, yielded higher accuracy. It is not yet clear whether this difference arises from dataset size or from variations in fruitlet growth caused by differing environmental conditions between years. Future analyses should investigate the impact of larger and more diverse training datasets on performance. As additional growing seasons are recorded and more data becomes available, we expect further improvements in robustness and generalization across varietals and conditions.

Our current approach is designed for image capture followed by offline processing, where images are captured in the field and processed later on more powerful computational hardware. This design choice is well-suited to the practical requirements of thinning prediction, as growth rate calculations inherently require measurements across multiple days and do not need to be computed immediately at the time of image capture. This setup allows us to leverage more computationally demanding models, such as the two-stage Mask-RCNN for segmentation and deep-learning-based disparity estimation methods like RAFT-Stereo. However, this workflow does require connecting the camera to a device with sufficient memory capacity to store the captured images for later processing.

If real-time sizing were desired in the field, a computationally sufficient device would need to be carried and remain connected to the camera during data collection. This brings challenges as the wiring makes it difficult to maneuver around the cluster. In an ideal scenario, lightweight models small enough to run on mobile devices would be used. While it is possible to replace the backbone of Mask-RCNN with lightweight networks \cite{mobilenet, mobile_mask}, RAFT-Stereo still requires sufficient computational resources. Further investigation is needed to determine if faster learning-based stereo matching algorithms \cite{realtime_stereo, hitnet} provide sufficient performance to be used in agriculture.

Lastly, one drawback of our approach is that the system cannot distinguish between a fruitlet that fell off and a missed detection. This affects thinner response predictions as the Fruitlet Growth Model considers fruitlets that have abscised during the sizing period. Future work would have to be dedicated towards this in order to fully replicate the process described in the model.

\subsection{Future Directions for Autonomous Deployment}
The next step to make the process fully autonomous is to replace the need for humans to collect images. This paper evaluates the sizing performance of images collected by a robot following a set motion path. The system fails to successfully size smaller, fully occluded fruitlets and large, partially occluded fruitlets. This is because the robot is not reasoning about the environment as it captures images, unlike a human operator with a handheld camera. However, an additional advantage of our computer vision system is that it can be integrated into more intelligent next-best-view planning pipelines~\cite{freeman_nbv}.

In future work, we focus on integrating our computer vision and growth rate tracking pipeline into a fully autonomous robotic system that can drive down a row of trees, identify AprilTags, intelligently take pictures, size, and associate the corresponding clusters with no human intervention. Previous works have been dedicated towards autonomously navigating rows of crops \cite{silwal2021bumblebee}. We believe such a system has the potential to significantly reduce the labor costs required to size apple fruitlets, and ultimately enable growers to optimize yield.
\vspace{-5pt}
\section{Conclusion}\label{sec:conclusion}

This work presents an alternative approach to sizing and measuring the growth rates of apple fruitlets. 
We evaluated our method on data collected across two growing seasons and three apple varietals. Our computer vision-based system produces fruitlet size and growth rate measurements comparable to those obtained using hand calipers and predicts abscission rates within 0.4–7.4\% of traditional measurements. These results were achieved using only one to two stereo image pairs per cluster, without any human effort required to label fruitlets or take caliper measurements. Through ablation studies, we analyzed the contribution of individual components and developed a modular pipeline in which detection, stereo, and association modules can be replaced as new state-of-the-art methods emerge.

Future work will focus on improving robotic data collection through next-best-view planning and advancing towards a fully autonomous system capable of navigating orchard rows, detecting AprilTags, capturing images, and performing complete end-to-end sizing and growth rate estimation to enable large-scale adoption in commercial orchards.

\section*{Acknowledgments}
We would like to thank the University of Massachusetts Amherst Cold Spring Orchard for allowing us to collect data. This research was funded by NSF / USDA NIFA 2020-01469-1022394 and NSF Robust Intelligence 195616.

\section*{Declaration of generative AI and AI-assisted technologies in the manuscript preparation process}
During the preparation of this work the authors used ChatGPT (OpenAI, GPT-5 model) in order to assist with editing and improving the clarity of the manuscript. After using this tool, the authors reviewed and edited the content as needed and take full responsibility for the content of the published article.

\printcredits
\clearpage

\bibliographystyle{cas-model2-names}

\bibliography{mybib}





\end{document}